\documentclass[journal]{IEEEtran}

\usepackage{cite}
\usepackage{amsmath,amssymb,amsfonts}
\usepackage{graphicx}
\usepackage{textcomp}
\usepackage{subfig}
\usepackage[flushleft]{threeparttable} 
\usepackage{booktabs}
\usepackage{algorithm} 
\usepackage{algpseudocode}
\usepackage[colorlinks = true,
            linkcolor = blue,
            urlcolor  = blue,
            citecolor = blue,
            anchorcolor = blue]{hyperref}
\usepackage{microtype}
\allowdisplaybreaks

\usepackage{url}

\usepackage{comment}

\usepackage{multicol}
\usepackage{multirow}
\usepackage{dsfont}
\usepackage{mathtools}
\usepackage{cuted}
\usepackage[makeroom]{cancel}
\usepackage{stmaryrd}
\usepackage{booktabs, makecell}
\usepackage{pifont}
\usepackage{ulem}
\usepackage{color}
\definecolor{b2}{RGB}{51,153,255}
\definecolor{p2}{RGB}{121,64,255}
\definecolor{y2}{RGB}{5, 131, 123 }
\definecolor{r2}{RGB}{233, 84, 24}

\newcommand{\Peng}[1]{{\color{red}[Peng: #1]}}
\newcommand{\Li}[1]{{\color{p2}[Li: #1]}}
\newcommand{\Wang}[1]{{\color{cyan}[Wang: #1]}}
\newcommand{\WP}[1]{{\color{black}[Wang \& Peng: #1]}}
\newcommand{\ZL}[1]{{\color{p2}[Zhu \& Li: #1]}}

\newcommand{\appremove}[1]{}

\newcommand{\app}[1]{#1}
\newcommand{\revision}[1]{#1}
\newcommand{\revisionfoot}[1]{}

\newcommand{\ie}{{\it{i.e.,~}}}
\newcommand{\eg}{{\it{e.g.,~}}}

\newcommand{\etal}{{\it{ et al.~}}}     

\newcommand{\R}{\mathcal{R}}

\newcommand{\ours}{\textbf{\texttt{FedNI}}}
\newcommand{\MNG}{missing node generator~}

\newcommand{\cm} {\color{red}}
\newcommand{\eat}[1]{}

\def\BibTeX{{\rm B\kern-.05em{\sc i\kern-.025em b}\kern-.08em
    T\kern-.1667em\lower.7ex\hbox{E}\kern-.125emX}}

\begin{document}

\title{FedNI: \textbf{Fed}erated Graph Learning with \\ \textbf{N}etwork \textbf{I}npainting for Population-Based \\ Disease Prediction }

\author{Liang Peng, Nan Wang, Nicha Dvornek, Xiaofeng Zhu, and Xiaoxiao Li \IEEEmembership{Member, IEEE}
\thanks{L. Peng and N. Wang are equally contributed to this work.}
\thanks{L. Peng and X. Zhu are with the Center for Future Media and School of Computer Science and Technology, University of Electronic Science and Technology of China, Chengdu 611731, China.}
\thanks{N. Wang is with School of Computer Science and Technology, East China Normal University, Shanghai, 200062, China, and the University of British Columbia, Vancouver, BC V6T 1Z4  Canada}
\thanks{N. Dvornek is with the Department of Radiology and Biomedical Imaging, Yale University, New Haven, CT 06511 USA }
\thanks{X. Li is with Electrical and Computer Engineering, the University of British Columbia, Vancouver, BC V6T 1Z4  Canada (e-mail:xiaoxiao.li@ece.ubc.ca).}
}
\maketitle

\begin{abstract}

Graph Convolutional Neural Networks (GCNs) are widely used for graph analysis. Specifically, in medical  applications, GCNs can be used for disease prediction on a population graph, where graph nodes represent individuals and edges represent individual similarities. However, GCNs rely on a vast amount of data, which is challenging to collect for a single medical institution. In addition, a critical challenge that most medical institutions continue to face is addressing disease prediction in isolation with incomplete data information. To address these issues, Federated Learning (FL) allows isolated local institutions to collaboratively train a global model without data sharing. In this work, we propose a framework, \ours, to leverage network inpainting and inter-institutional data via FL. Specifically, we first federatively train missing node and edge predictor using a graph generative adversarial network (GAN) to complete the missing information of local networks. Then we train a global GCN node classifier across institutions using a federated graph learning platform. The novel design enables us to build more accurate machine learning models by leveraging federated learning and also graph learning approaches. We demonstrate that our federated model outperforms local and baseline FL methods with significant margins on two public neuroimaging datasets. 

\end{abstract}


\section{Introduction} \label{sec_intro}


 Neurological disorders and diseases, such as autism spectrum disorder (ASD) and Alzheimer’s disease (AD) \cite{stevens2021ankyrins}, can cause significant social, communication, cognitive, and behavioral challenges \cite{wee2016diagnosis,song2020classification}.
 It is highly desired to detect neurological disorders, promoting early intervention and effective treatment of the disease in clinics. 
Recent studies have applied deep learning techniques for early disease diagnosis \cite{huang2020edge,song2019graph}, 
such as convolutional neural networks (CNNs) \cite{li2018brain,spasov2019parameter}, recurrent neural networks (RNNs) \cite{dvornek2019jointly,sudha2021recurrrent}, and graph convolutional neural networks (GCNs) \cite{song2020classification,parisot2018disease}. 
Although CNNs and RNNs have achieved plausible and promising results in the early diagnosis of ASD and AD, they extract individual imaging information independently and have limitations in exploring information from unlabeled individuals and the complex structure inherent to the data, \eg ignoring the interaction and association between subjects in the population, failing to guarantee learning effective models.

Disease prediction on a population can be naturally modeled using graphs or networks\footnote{Graph and network are interchangeable in our manuscript.}. Specifically, the nodes are represented as image features and labeled by their health condition (patients or healthy controls), while the edge {that connect two nodes} capture the similarity between individuals.  GCNs can synergize the representation power of all kinds of information on individual subjects to predict individual labels based on partially labeled individual subjects together with the interactions among the whole population. Hence, GCNs are widely used for graph analysis in disease prediction. 
For example, Parisot \etal  applies GCN for semi-supervised  disease prediction on neuroimaging data, where nodes are defined as subjects and an edge  represents the  interaction and association between two subjects \cite{parisot2018disease}.


A critical challenge that most medical institutions continue to face is addressing disease prediction in isolation without any insight from other institutions. Specifically, multi-institutional collaboration that centrally shares patient data faces privacy and ownership challenges, such as  general data protection regulation (GDPR) \cite{voigt2017eu} and health insurance portability and accountability act (HIPAA) \cite{act1996health}. 
To address this issue, Federated Learning (FL) allows isolated institutions to collaboratively `utilize' their private data without data sharing but transmit knowledge to each other.
For example, Li \etal \cite{li2020multi} proposes to implement a decentralized iterative optimization algorithm and preserve the privacy of shared local model weights through differential privacy for disease diagnosis. Yang \etal~\cite{dayan2021federated} 
proposes to share only a partial model between the server and institutions to automatically conduct diagnosis of COVID-19.

However, most existing FL strategies were designed for CNNs and multilayer perceptrons (MLPs) \cite{yang2021flop,astillo2021federated}, while few methods have learned  to combine GCN with FL for disease prediction \cite{he2021fedgraphnn}. Different from CNNs and MLPs with only the feature matrix as the input, GCN has another input, \ie the graph containing the interaction and association information between two subjects. If implementing GCN in FL, the global graph will be separated into several local graphs, where edges between local graphs are missing. When focusing on a node with missing neighbors (an ego), the corrupted information destroys its original ego network structure\footnote{An ego network is the graph of all nodes that are less than a certain distance, such as  immediate neighbors, from a focal node (``ego'') \cite{hanneman2005introduction}.}. Ego-network information has shown to be important in GCN-based node classification~\cite{zhang2021subgraph, zhu2020transfer}. However, the current GCN models overwhelmingly assume that the node and edge information are complete. As a result, simply combining GCN 
with the standard cross-silo FL strategy on distributed local graphs could undermine the effectiveness of GCN. \revision{One solution is to connect the local graphs,but it requires sharing node information across clients and may violate privacy regulations. Therefore,}\revisionfoot{\footnote{Reviewer 2, Q2}} to address this issue, network inpainting, which predicts the missing neighbors and their associated edges, is a promising strategy to improve the completeness of the graph for GCN learning~\cite{taguchi2021graph}.

\begin{figure*}[t]
\centering
 \includegraphics[scale=0.395]{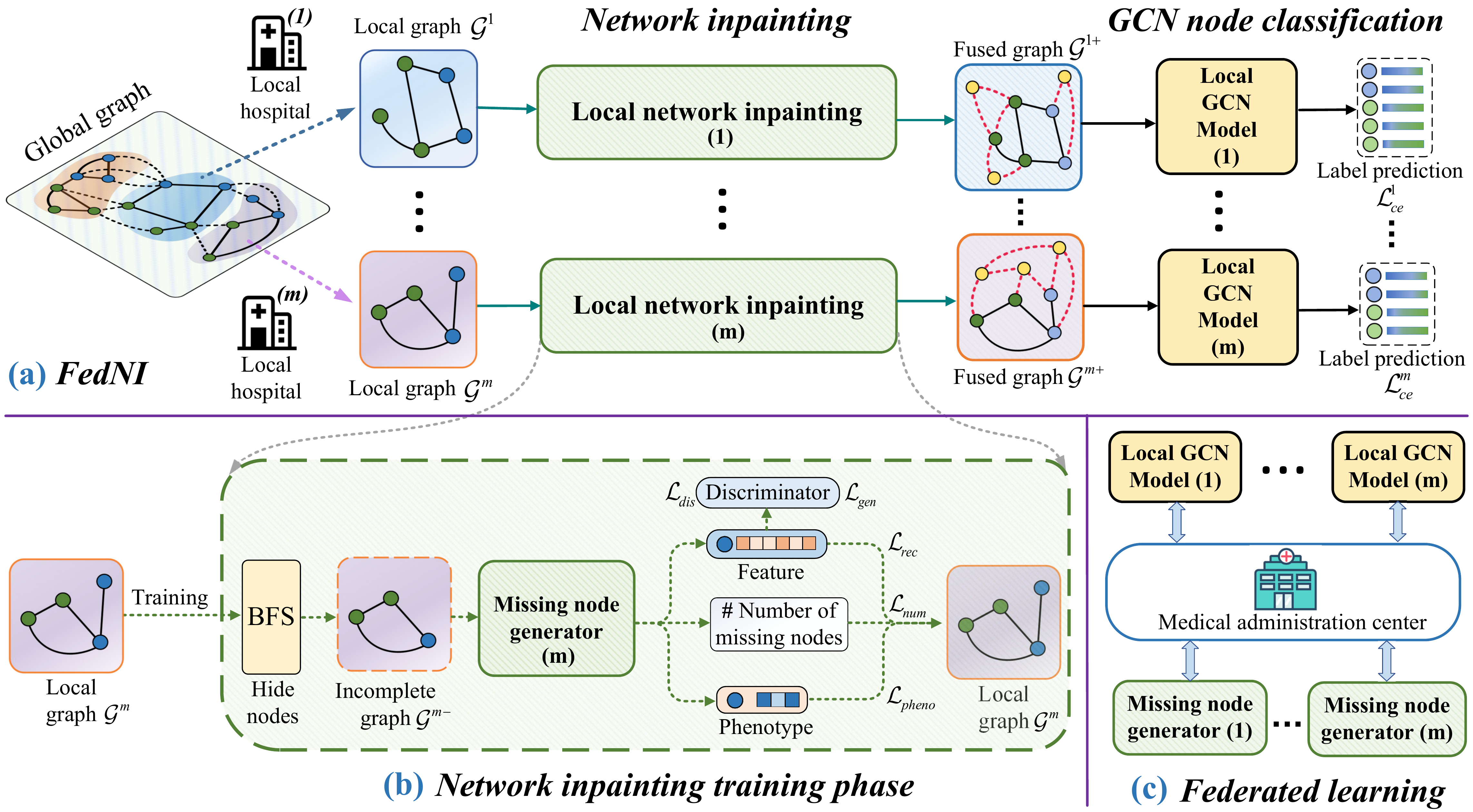}
		\caption{The flowchart of the proposed method.  Specifically, given $m$ local hospitals and a medical administration, we construct $m$ local population graphs containing nodes (individual subjects), node features (imaging representations), and edges (similarities). The links between the local nodes and the unseen nodes in the other hospitals are missing (dash links in the global graph). The green block shows SN-GAN-based network inpainting module to predict missing nodes and generate a fused graph, which is trained first. Then, we generate fused graphs from the pretrained generator from the original local graphs. Last, the fused graphs are fed into the yellow block, a GCN node prediction module, to predict node labels. The missing node generators in the local network inpainting modules and the local GCN model for node classification are trained using a FedAvg~\cite{mcmahan2017communication} scheme coordinated by a trusted medical administration center. The detailed procedure of our proposed framework is in Algorithm \ref{alg:fedgdp:main}}
\label{fig:flowchat}
\end{figure*}

In this work, we propose a new FL framework for disease prediction, \ours, which conducts the GCN node classification on a population graph and addresses the above issues. To this end, we propose a two-phase FL pipeline (as shown in Fig. \ref{fig:flowchat}). In the first phase, we start with training missing node generators for network inpainting. Specifically, we randomly remove nodes from the local graph with Breadth-First Search (BFS), resulting in a set of corrupted graph and \revision{hidden nodes}\revisionfoot{\footnote{Reviewer 1, Q6}}. We train the generator to predict the hidden nodes and edges from the corrupted graphs using spectral normalized generative adversarial networks~(SN-GAN)~\cite{miyato2018spectral}. Then, we feed the original local graphs to the trained missing node generator to inpaint missing neighbors and generate new fused graph with augmented nodes and edges. In the second phase, we federatively train GCN for node classification on local fused graph, where each institution iteratively trains local GCN model weights, shares them with a global trusted administration center, and then receives the averaged model weights.

Our main contributions are summarized as follows:
\begin{enumerate}
    \item We are the first to formulate cross-silo federated graph learning for disease classification using disjointed small local graphs that are partitioned from a global population graph. We increase the effectiveness of GCN on ego-networks by a novel network inpainting module and improve performance by training node generators and node classifiers in a FL framework.

    \item The proposed FL network inpainting module 
    includes \textit{three components} to ensure its effectiveness:  BFS node removal to avoid generating isolated corrupted local graphs; SN-GAN to synthesize nodes with realistic data distribution; and edge construction employing both image and non-image information to better integrate information over the graph. 
    \item We conduct extensive experiments on comparisons, ablation study and hyperparameter discussion. 
    Our method achieves $66.7\%$ and $75.8\%$ classification accuracy on the widely-used ABIDE and ADNI datasets, respectively,  notably outperforming the alternative FL methods.
\end{enumerate}
\section{Related work}


\subsection{ GCNs for disease prediction}

Recently, there has been an increasing focus on graph learning on unstructured data in the medical domain. 
For example, Parisot~\etal~\cite{parisot2018disease} exploits the GCN and involves representing populations as a sparse graph in which nodes are associated with imaging features, and phenotypic information is described as edge weights. 
Hi-GCN~\cite{jiang2020hi} uses a hierarchical GCN to learn the graph feature embedding for the classification of ASD and AD.

However, two issues remain in the current GCN models. 
First, the effectiveness of GCN highly relies on the quality of the graph. When a low-quality graph is input, \ie a graph with missing component~\cite{taguchi2021graph, zhang2021subgraph}, the classification performance is affected. To address the problem of graphs containing missing features, the most popular strategy is to estimate and fill in the unknown values before~\cite{yi2019not, luo2018multivariate} or jointly with ~\cite{zhang2021subgraph} applying GCN. Different from the existing methods, we combine edge augmentation strategies to generate fused graphs.
Second, many GCNs solely rely on a single modality, \ie imaging data \cite{abraham2017deriving}, to explore the similarity between two nodes. Consequently, they fail to comprehensively capture interactions and similarities between subjects or their individual scans~\cite{zhu2022interpretable}. To tackle this issue, we explore imaging and non-imaging data to represent the population graph. 


\subsection{Federated learning for disease prediction}
Federated learning (FL)~\cite{mcmahan2017communication} is a collaborative and decentralized privacy protection technology, which is designed to tackle the problem of data island while protecting data privacy \cite{li2020review}. 
 Driven by the achievement of FL in data privacy protection, FL could have wide popularization and application prospects in data-sensitive fields, such as disease prediction. 
Li~\etal~\cite{li2019privacy} is one of the first works that implements a practical FL system for brain tumor segmentation. Li \etal \cite{li2020multi} apply FL and domain adaptation techniques on heterogeneous multi-institutional neuroimaging data analysis. Dayan \etal~\cite{dayan2021federated} promote COVID patient risks prediction via collaborating 20 hospitals in FL.


In contrast to the previous FL research, we investigate an under-explored but realistic setting of expressing multi-centric medical data as distributed population graphs and address the unique challenge of FL on a local graph with missing neighbors. The closest related work is Zhang \etal~\cite{zhang2021subgraph}. However, this work targets the general graph node classification problem and does not consider the uniqueness of our considered disease population network (\ie small local subgraphs), utilize auxiliary information (\eg phenotypes), or consider the extreme privacy requirement in healthcare applications. Our proposed \ours~ improves upon Zhang \etal~\cite{zhang2021subgraph} by addressing the above limitations.

\section{The Proposed Method}
\label{sec_method}

The goal of our work is to enable multiple institutions to collaboratively learn a powerful GCN model for population-based disease prediction without data sharing. For this purpose, we first define the population graph formulation (in Sec.~\ref{sec_A}). Then, we propose a two-phase FL framework as shown in Fig.~\ref{fig:flowchat}, which contains a network inpainting module with missing node generator and a GCN learning module for node classification. The local network inpainting module (in Sec.~\ref{sec_mng}) is designed to predict missing nodes and augment edge connections, resulting in a larger fused local graph that serves as the input to the GCN node classifier. To leverage the distributed data, the two proposed modules are trained in a certain FL fashion (in Sec.~\ref{sec_fl}). 

\subsection{Population graph formulation} \label{sec_A}



Graph data consists of a set of nodes and a set of edges. Each node is represented by the node feature containing the properties of this node, while each edge implies the similarity between two nodes.
A standard approach to construct graph edges is performing clustering based on node features. However, this approach has two limitations. First, if node features are high dimensional vectors, it is difficult to reveal the real local structure due to the curse of dimensionality. Second, generating edges using a single modality is not enough to mine the complex relationship inherent to the graph data.
To address the above challenges, inspired by Parisot \etal \cite{parisot2018disease}, in this paper, we generate edges by conducting dimensionality reduction  on the image features as well as considering the phenotypic information of the subject. 


First, we derive low-dimensional and discriminative features from raw medical images and then use them to construct a similarity graph $\mathbf{S} \in \mathbb{R} ^{n \times n}$ where $n$ denotes the number of nodes in the population graph, aiming at reducing the adverse influence of high-dimensional features, \eg noisy and redundant features as well as the curse of dimensionality.
Specifically, given the raw feature matrix $\mathbf{X} \in \mathbb{R} ^{n \times d}$,  we employ the Gaussian kernel to define the weight of the edge as:
\begin{equation}   
\mathbf{s}_{ij} = \mathrm{exp}\left(-\frac{ \left ( \mathbf{h}_i-\mathbf {h}_j\right)^2 }{2 \sigma ^{2}} \right),
\label{Eq5}
\end{equation}
where $\mathbf{H} \in \mathbb{R} ^{n \times d_h} (d_h < d)$ is the low-dimensional features extracted by dimensionality reduction algorithms (\eg principal component analysis and recursive feature elimination) and $\sigma$ is the width of the Gaussian kernel.

Second, we calculate node similarity from another angle by utilizing phenotype data (\eg sex, age and gene), aiming at providing much information to output high-quality graphs.  
Specifically, given the phenotype data matrix $\mathbf{U}^{q} \in \mathbb{R} ^{n \times d_q}$ where $q$ ($q = 1, ..., Q$) denotes different types of phenotype data, we define a phenotypic graph $\widetilde{\mathbf{S}}$ as 
\begin{equation}   
\widetilde{\mathbf{s}}_{ij}  =  \sum_{q=1}^{Q}  f^{q}(\mathbf{u}^{q}_{i},\mathbf{u}^{q}_{j}),
\label{Eq6}
\end{equation}
where $Q$ represents the number of different phenotypes and $f^{q}(\cdot)$ is the similarity measure function for the $q$-th phenotype data. As the phenotype representation's distribution and underlying generation process are different from node feature, we can pick different similarity metrics from that of node features in Eq.~~\ref{Eq5}. Particularly in this work, we define $f^q$ as  $\mathbb{I}\{\mathbf{u}^{q}_{i}=\mathbf{u}^{q}_{j}\}$ for sex and gene phenotype data, and $\mathbb{I}\{\left \| \mathbf{u}^{q}_{i}-\mathbf{u}^{q}_{j} \right \| \le \gamma \}$ where $\gamma > 0$  for age phenotype data.

Finally, we fuse the edges derived from image-based node features (Eq.~\ref{Eq5}) and the edges derived from phenotype information (Eq.~\ref{Eq6}) to obtain the initial graph $\mathbf{A}$ by performing the Hadamard product between the similarity graph matrix $\mathbf{S}$ and the phenotypic graph matrix $\widetilde{\mathbf{S}}$ (\ie $\mathbf{A} =  \mathbf{S} \circ \widetilde{\mathbf{S}}$). 
Moreover, we sparse graph $\mathbf{A}$ by keeping $k$ edges with the largest weights for each node and setting others as zeros.
Furthermore, we add the diagonal matrix $\mathbf{I}$ to $\mathbf{A}$ (\ie $\mathbf{I} +  \mathbf{A} \rightarrow \mathbf{A}$). 


\subsection{Network inpainting}  \label{sec_mng}

 In FL,  each local graph is a subset of the global graph and there is no overlapped nodes across the local graphs of different institutions. Thus, if the neighbors of local graph nodes are located in another institution, the local graph is incomplete. Focusing on a certain node with missing neighbors, its original ego network structure of the global graph is incomplete. Thus, the node information of local graph is incomplete. Directly apply GCNs on the corrupted graph with missing features may result in degraded and unstable performance.
To address this, we first design a missing node generator to generate missing nodes and edges for each local graph. 
We then conduct FL on local missing node generators to generate new nodes and edges to complete the ego-networks of the corrupted local graph and augment message passing. 

\subsubsection{Node masking for missing node generation}\revisionfoot{\footnote{Reviewer 2, Q1}} \label{sec_bfs}



A practical approach to predict missing nodes for graph inpainting is training a self-supervised regression model to predict the hidden nodes and then using this model to predict missing node in the local graphs. 
To accomplish this, we need to hide nodes and edges of each local graph to construct the regression model. 
In the literature,  the random method is widely used to randomly remove nodes~\cite{zhang2021subgraph}, but it is not conducive for feature propagation in the graph. Moreover, the random method may generate isolated and disconnected components
that may hurt the completeness of the ego-network structure, resulting in sub-optimization of the generation network.
To overcome the above issues, we propose to use the BFS algorithm to represent a graph as a subtree structure for a given root node.
The intuition is that removing the leaf nodes in the subtree structure has little effect on applying our selected  Weisfeiler-Lehman algorithm GCN strategy~\cite{kipf2016semi} to the local graph structure.

Specifically, given a connected graph $\mathcal{G} =\{V,E\}$ with a root node $v_r \in V$, we first build a BFS tree by traversing the nodes in the graph with information broadcasts to neighbor nodes.  
Based on this property, we can generate sub-graphs by removing leaf nodes at different depths. 
More specifically, selecting a root node ($v_{r}$) and using the BFS algorithm on the local graph $\mathcal{G}$, we mask some nodes belonging to the $b$-th depth to obtain an incomplete local graph $\mathcal{G}^{-}$ as
\begin{eqnarray}
\mathcal{G}^{-} = f_{\mathrm{bfs}}(\mathcal{G}; v_r; b).
\label{eq3}
\end{eqnarray}
By randomly visiting the nodes of the local graph $\mathcal{G}$ as root nodes and removing its leaf nodes, we can generate many pairs (incomplete graph $\mathcal{G}^{-}$, hidden nodes) that are used as (inputs, labels) for missing node generator training (\revision{the number of hidden nodes usually take up 10\%-15\% of the total nodes in our experiments}\revisionfoot{\footnote{Reviewer 3, Q7}}). 
We further denote $v_{k} \in \mathcal{N}_{i}^{-}$ as one of masked neighbors of the node $v_{i}$ ($v_{i} \in  \mathcal{G}^{-}$), \revision{and $\mathcal{N}_{i}^{-}$ denote the set of missing neighbors of $v_{i}$. As each masked node has its corresponding parent node in the graph $\mathcal{G}^{-}$, the goal of the missing generator is to reconstruct the information of masked neighbors $v_{k}$ by its parent node $v_{i}$ based on the incomplete local graph $\mathcal{G}^{-}$.}

\subsubsection{Local network inpainting} \label{sec_gan}

After masking some nodes by the BFS algorithm in each local graph,  we need to correctly predict the number of missing neighbors for each node as well as the corresponding feature and phenotype information of each missing neighbor (\ie missing node).

As GCN \cite{kipf2016semi} and its variants (\eg GraphSAGE \cite{hamilton2017inductive}) have been widely used to  capture semantic information and structural information in the graph data, in this study,  we employ  a GCN encoder $f_{\theta _{g}}(\cdot)$  to obtain the embeddings of each node in the local graph $\mathcal{G}^{-}$. 
Note that GCN is more robust for small graph data, while GraphSAGE is more flexible for large-scale graph data~\cite{wu2020comprehensive}. Therefore, we apply GCN in this study considering that the local population graph of a local system is usually a small graph.
\revision{Although different GNN structures can result different performance, our primary focus in this paper is not on different GNN structures, but on federal learning over population graph. Therefore we choose the most commonly used GNN structure (\ie GCN).}\revisionfoot{\footnote{AE, Q2 and Reviewer 3, Q1}}
Concretely, with $\mathcal{G}^{-} = \{ \mathbf{X}^{-}, \mathbf{A}^{-}\}$ \revision{where $\mathbf{X}^{-}$ and $\mathbf{A}^{-}$ means the feature matrix of nodes and the graph matrix of the incomplete graph}\revisionfoot{\footnote{Reviewer 3, Q6}}, the embeddings are obtained by 
\begin{equation}   
\mathbf{Z} = f_{\theta _{g}}( \mathbf{X}^{-}, \mathbf{A}^{-}),
\label{eq8}
\end{equation}
where the GCN encoder $f_{\theta _{g}}$ has several hidden layers, and each hidden layer includes  two operations, \ie feature learning and neighborhood aggregation. 
More specifically, the GCN operation on the $l$-th hidden GCN layer is defined as:
\begin{eqnarray}
f_{\theta _{g}}^{(l)}( \mathbf{Z}^{(l)}, \mathbf{A})=\sigma (\mathbf{D}^{-\frac{1}{2}}\mathbf{A}\mathbf{D}^{-\frac{1}{2}}\mathbf{Z}^{(l)}\mathbf{\Theta}_{g}^{(l)}),
\label{eqgcn}
\end{eqnarray}
where $\mathbf{D}$ is the diagonal matrix of $\mathbf{A}$, and $\mathbf{\Theta}_{g}^{(l)}$ is a weight matrix which needs to be trained in the $l$-th layer, and $\sigma(\cdot)$ represents the function for activation operation. 

Furthermore, our proposed missing node generator constructs a regression model between the embeddings $\mathbf{Z}\in \R^{n \times d_z}$ and the ground truth to predict the number of missing nodes/neighbors, the features and the phenotype information of these nodes.

\noindent  \textbf{Predicting the number of missing nodes.}
The number of missing nodes for each parent node is predicted by 
\begin{equation}   
\mathcal{L}_{num}= \sum_{i \in \mathcal{V}^{m}} \left \| N_{i}^{-}- \mathcal{S}(f_{\theta _{n}}(\mathbf{z}_{i}))   \right \| ^{2}_{2},
\label{eq15}
\end{equation}
where $N_{i}^{-}$ denotes the number of missing neighbors of the $i$-th node which is normalized to $[0,1]$, $f_{\theta_{n}}(\cdot)$ is a MLP predictor mapping the embedding $\mathbf{z}_i \in \R^{d_z}$ to an integer, $\mathcal{S}(\cdot)$ is the sigmoid function, and $\mathbf{z}_{i}$ is the embedding of $v_{i}$ obtained by Eq. (\ref{eq8}). 
 Since the embedding $\mathbf{z}_{i}$ aggregates the information from its remaining neighbors by the GCN encoder, 
the predictor $f_{\theta_{n}}(\cdot)$ can be used to predict various number of missing neighbors based on the different representations of $\mathbf{z}_{i}$ for $v_i \in \mathcal{G}^{-}$. 


\noindent  \textbf{Node feature prediction.}
Jointly with predicting the number of missing nodes, we predict the feature and the phenotype information of each missing node.
Specifically, we design another regression model (\ie a MLP), $f_{\theta_{z}}(\cdot)$, taking in the embedded feature $\mathbf{z}_{i}$ form the GCN encoder $f_{\theta _{g}}(\cdot)$ to reconstruct the feature of each masked neighbor (\eg $v_{k} \in \mathcal{V}^{-}, k \in \mathcal{N}_{i}^{-}$) from its parent node $v_{i}$ by
\begin{equation}   
\tilde{\mathbf{x}}_{k}= f_{\theta_{z}}( \mathcal{R}(\mathbf{z}^{-}_{i})),
\label{eq9}
\end{equation}
where $\mathcal{R}(\cdot)$ is a Gaussian noise generator for generating diverse missing neighbors.
Moreover, we introduce the reconstruction loss $\mathcal{L}_{rec}$ to optimize the parameters  $\theta _{g}$ and $\theta_{z}$ by 
\begin{equation}   
\mathcal{L}_{rec}= \sum_{i \in \mathcal{V}^{m}}\sum_{k \in \mathcal{N}_{i}^{-}} \left \| \mathbf{x}_{k}- \tilde{\mathbf{x}}_{k} \right \| ^{2}_{2}.
\label{eq10}
\end{equation}

The missing node generator can be regarded as an encoder-decoder. That is,
$f_{\theta _{g}}(\cdot)$ is the GCN encoder which considers the incomplete graph information (\ie $\mathcal{G}^{-}$), while $f_{\theta_{z}}(\cdot)$ is the MLP decoder for reconstructing the features of  masked neighbors. 
However, it is challenging to make the features generated by the generator realistically match the data distribution of the missing neighbors. To do this, we adopt the SN-GAN  \cite{miyato2018spectral} to improve the effectiveness of the generator. 
Specifically, letting $\mathcal{X}$ be the original data distribution and  $\widetilde{\mathcal{X}}$ be the generated data distribution, we obtain  the generator loss by
\begin{equation}   
\mathcal{L}_{gen}=-\mathbb{E}_{\tilde{\mathbf{x}}\sim \widetilde{\mathcal{X}} }\left[1- \mathrm{log}\left(f_{\theta _{d}}(\mathbf{\tilde{x}}) \right)  \right] ,
\label{eq11}
\end{equation}
where $f_{\theta _{d}}(\cdot)$ denotes the discriminator to evaluate the gap between generated features and the features of hidden neighbors.  Integrating reconstruction loss (Eq. \ref{eq10}) with generator loss (Eq. \ref{eq11}), the final loss for feature reconstruction is formulated as:
\begin{equation}   
\mathcal{L}_{fea}= \alpha \mathcal{L}_{rec} + \beta \mathcal{L}_{gen} ,
\label{eq12}
\end{equation}
where $\alpha$ and $\beta$ are the weights of $\mathcal{L}_{rec}$ and $ \mathcal{L}_{gen}$, respectively.
The discriminator $f_{\theta _{d}}(\cdot)$  is trained to identify if the feature comes from the generator or the real missing neighbor. 
\begin{equation}   
\mathcal{L}_{dis}= -\underset{\mathbf{x}\sim \mathcal{X} }{\mathbb{E}} \left [1- \mathrm{log}\left(f_{\theta _{d}}\left(\mathbf{x}\right)\right )  \right ]-\underset{\tilde{\mathbf{x}}\sim \widetilde{\mathcal{X}} }{\mathbb{E}} \left [\mathrm{log}(f_{\theta _{d}}(\tilde{\mathbf{x}}))  \right ].
\label{eq13}
\end{equation}
In sum, the features of  missing nodes are reconstructed by node feature predictor  $f_{\theta_{z}}(\cdot)$ in Eq.~(\ref{eq9}), which is optimized by reconstruction loss (Eq.~(\ref{eq10})) and GAN loss (Eq.~(\ref{eq11}) and Eq.~(\ref{eq13})). 

As the features of missing nodes may not be enough to explore the complex structure across subjects in a disease population network, we further predict the phenotype to obtain the connections (\ie edges) for missing nodes.

\noindent  \textbf{Phenotype prediction and edge construction.} \label{sec_phe}
Given the number of missing nodes as well as their features, we use both the phenotype information of these nodes and their node property to find the connection between each missing node and known nodes.
In this respect, we first adopt a model $f_{\theta_{u}}(\cdot)$ to predict the phenotype information for masked nodes and then find the connection between two nodes.

For the $k$-th masked neighbor of the parent node $v_{i}$, the $q$-th type of phenotype prediction $\tilde{\mathbf{u}}_{k}^{q}$ is generated by $\tilde{\mathbf{u}}_{k}^{q} = f_{\theta_{u}}(\mathcal{R}(\mathbf{z}^{-}_{i}))$, where $\mathbf{z}^{-}_{i}$ and $\mathcal{R}(\cdot)$  are defined as in Eq. (\ref{eq9}). 
For phenotype data with class labels (\eg sex and gene),  we adopt the cross-entropy loss function to update model parameters
\begin{equation}   
\small
\mathcal{L}_{pheno} \!=\! \sum_{i \in \mathcal{V}^{}}\sum_{k \in \mathcal{N}_{i}^{-}} \left ( \mathbf{u}_{k}^{q}\log(\tilde{\mathbf{u}}_{k}^{q}) \!+\! (1-\mathbf{u}_{k}^{q})\log(1-\tilde{\mathbf{u}}_{k}^{q}) \right ) ,
\label{eq14}
\end{equation}
where $\mathbf{u}_{k}^{q}$ is the phenotype label. Similarly, age prediction can be optimized by regression losses.

 It is worth noting that our \ours's local graph inpainting strategy significantly differs from that used by Zhang \etal~\cite{zhang2021subgraph}. The differences are as follows: 1) We improve random node masking to BFS-based node masking to avoid generating isolated components in local graph; 2) We incoperate SN-GAN-based training scheme and novel loss functions to improve the quality of generated node features, instead of requiring auxiliary information from the other local graphs that can expose more privacy risk; 3) In addition to predicting missing node feature, we also predict its associated phenotype to construct weighted edges; 4) The phenotype is further used to rebuild new edges to augment message passing. The advantages of our strategies are verified in Sec.~\ref{sec_ablation}. 
 
After obtaining the phenotype information of each missing node, we can generate edges for these missing nodes by the same way as described in Sec. \ref{sec_A}. 

\subsection{Federated learning}
\label{sec_fl}
Although the \MNG is able to generate missing nodes for each local graph, it is well known that GAN may perform poorly on small data~\cite{zhao2020differentiable}. Similarly, training an effective GCN model for node classification also requires a large amount of samples. With the privacy constraints on analyzing multi-institutional medical data, the proposed \ours~ includes two FL phases: i) Federated network inpainting is proposed to fill in the incomplete features of local graphs, ii) Federated GCN is proposed to obtain a global node classifier. Our FL strategy follows the widely used FedAvg~\cite{mcmahan2017communication} aggregation strategy.  

\subsubsection{Federated network inpainting} \label{sec_fgel}
In Sec. \ref{sec_bfs}, we describe a local \MNG for each local system. To empower the node generator, we aim to federatedly learn a global node generator that leverages the local graphs from different institutions without centralizing the data. However, we keep the discriminator local as we notice averaging the discriminator can hurt model performance (shown in Sec~\ref{sec_ablation}). The intuition is that the generator should predict missing nodes following the data distribution of the global population network, while local discriminators can fit the heterogeneity of local graphs better. 
To achieve this goal, the FL process is to repeat the  following steps until convergence: i) each local \MNG (including the number of nodes predictor $f_{\theta _{n}}$ trained on Eq. (\ref{eq15}), node feature generator $f_{\theta _{g}}$ trained on Eq. (\ref{eq12}), and phenotype regression $f_{\theta _{u}}$ trained on Eq. (\ref{eq14})) is trained in parallel and updated; ii) the global \MNG  on the medical administration side aggregates and averages the local model parameters, and then broadcasts the updated model weights to all the local missing node generators. 

After the global \MNG is well trained and deployed to each local institution, we do \textit{graph merge} 
by performing network inpainting as inference on the original local graphs $\mathcal{G}^m$ in institution $m$, for $m\in [M]$. As described in Sec.~\ref{sec_mng}, our proposed \MNG predicts the number of missing neighbors $\tilde{n}_i$, missing node feature $\widetilde{\mathbf{x}}_i$, and their phenotype information $\widetilde{\mathbf{u}}_i$ for node $v_i$, for $v_i \in \mathcal{G}^m$. We further build new edges from the predicted nodes and existing nodes following the edge construction method illustrated in Sec.~\ref{sec_A}. We denote the new fused local graph of institution $m$ as  $\mathcal{G}^{m+} = (\mathbf{X}^{m+},\mathbf{A}^{m+}$).

\subsubsection{Federated GCN node classification} \label{sec_fgl}
As the ultimate goal of our method \ours~is to obtain a global GCN node classifier, we perform a federated GCN learning method to collaboratively train on  multiple local graphs while maintaining data privacy. After obtaining the fused local graphs $\mathcal{G}^{m+} (m \in [M])$ for all local institutions, we iteratively apply FedAVG~\cite{mcmahan2017communication} on GCN as follows: i) train each local GCN node classifier for a certain number of steps and then share GCN weights to the medical administration center, and ii) update the global GCN node classifier by averaging local GCNs' parameters and broadcast the updated global GCN to local institutions.

Specifically for the local GCN node classifier training, the $m$-th institution trains a local GCN node classifier $f_{\phi _{m}}(\cdot)$ with the fused graph  $\mathcal{G}^{m+}$ and its output is $\mathbf{P}^{m} = f_{\phi _{m}}(\mathbf{X}^{m+}, \mathbf{A}^{m+})$, where $f_{\phi _{m}}(\cdot)$ is expressed as Eq.~(\ref{eqgcn}) with learning parameters $\phi _{m}$.
Furthermore, the cross-entropy loss is regarded as the loss of the local GCN model: 
\begin{equation}   
\mathcal{L}_{ce}^{m}=-\sum_{i \in Y_{L}^{m}}^{} \left ( y_{i}^{m}log(p_{i}^{m}) + (1-y_{i}^{m})log(1-p_{i}^{m}) \right ), 
\label{eq16}
\end{equation}
where $Y_{L}^{m}$ is the set of labeled nodes on the $m$-th local system.
At the $t$-th iteration, the parameters $\phi _{m}^{t}$ of the $m$-th local model are updated (\ie $\phi _{m}^{t+1} \gets \phi _{m}^{t} - \eta \nabla \mathcal{L}_{ce}^{m}(\mathbf{X}^{m+};\mathbf{A}^{m+};\mathcal{Y}^{m})$) and sent to the server subsequently.

\subsubsection{Privacy preservation techniques} \label{sec_privacy_tech}
Recent work reveals that input data can be reconstructed from the shared model parameters~\cite{zhu2020deep}. To enhance the security of \ours, we employ differential privacy (DP)~\cite{dwork2014algorithmic}, which is a popular approach to privacy-preserving FL. For the parameters of local models, \ie missing node generator and GCN node classifier, shared from local institutions to the medical administration center, \revision{we add Gaussian noise with mean $0$ and standard deviation as $0.01$ to local model weights to satisfy differential privacy in this work.}\revisionfoot{\footnote{Reviewer 1, Q1}} 
\app{Above all, we present the detailed procedure of our proposed framework in Algorithm \ref{alg:fedgdp:main}.
\vspace{-1mm}
\begin{algorithm}
\footnotesize
\begin{algorithmic}[1] 
\app{
\State {\bf Require:} 
Data owners set $\{\mathcal{G}^1,\dots,\mathcal{G}^M\}$, server $S$, 
local network inpainting model with weights $\theta _{G}^{m} = \{ \theta _{g}^{m}, \theta _{z}^{m}, \theta _{n}^{m}, \theta _{h}^{m}\}$ for generator and $\theta _{d}^{m}$ for discriminator,
local GCN classifier $f_{\phi _{m}}(\cdot)$ with weights $\phi ^{m}$ on $m$-th client,
and the noise generator of differential privacy $\mathcal{T}(\cdot)$.

\State \textbf{First-phase}: Training missing node generators under FL setting (Run \textbf{Procedure A} and \textbf{Procedure C} iteratively)
\State \textbf{Obtain fused graphs}: $\{\mathcal{G}^{1+},\dots,\mathcal{G}^{M+}\}$\Comment{Network inpainting inference}
\State \textbf{Second-phase}: Training GCN node classification under FL setting (Run \textbf{Procedure B} and \textbf{Procedure D} iteratively) 

\State {\it On the server side:}
\Procedure{{\bf A } FederatedNetworkInpainting}{}
\For{$ t \leftarrow 1,2, \cdots,$} 
\State $\widetilde{\theta} _{G,t+1}^{m} \gets \textsc{LocalNetworkInpainting}(\mathcal{G}^{m+},\theta _{G,t}^{m}, \theta _{d,t}^{m}, t)$  
\State  $\theta _{G,t+1} \gets \frac{1}{M}\sum_{m\in[M]}\widetilde{\theta} _{G,t+1}^{m}$ and broadcast to local clients.
\EndFor
\EndProcedure
\Procedure{{\bf B }FederatedGCNNodeClassification}{}
\For{$ t \leftarrow 1,2, \cdots,$} 
\State Collect $\widetilde{\phi}_{t+1}^{m} \gets \textsc{LocalGCNClassifier}(\mathcal{G}^{m+},\phi _{t}^{m}, t)$  
\State $\phi_{t+1} \gets \frac{1}{M}\sum_{m\in[M]}\widetilde{\phi}^{m}_{t+1}$, and broadcast to local clients.
\EndFor
\EndProcedure
\State {\it On the data owners side:}
\Procedure{{\bf C} LocalNetworkInpainting}{$\mathcal{G}^{m} , \theta _{G}^{m} ,  \theta _{d}^{m} , t$}
\For{$ t_c \leftarrow 1,2, \cdots$} 
\State  $\{\mathcal{L}_{fea}, \mathcal{L}_{pheno}, \mathcal{L}_{num} \} $ $\leftarrow$ Eq. (\ref{eq12}), Eq. (\ref{eq14}), Eq. (\ref{eq15}) and back propagation. $\mathcal{L}_{dis}$ $\leftarrow$ Eq. (\ref{eq13}) and interval back propagation.
\EndFor
\State Send $\theta _{G,t+1}^{m} + \mathcal{T}(t+1)$ to Sever $S$
\State $\mathcal{G}^{m+} \gets \mathrm{Feedforward}(\mathcal{G}^{m}, \theta _{G}^{m})$ \Comment{Network inpainting inference} 
\EndProcedure

\Procedure{{\bf D }LocalGCNClassifier}{$\mathcal{G}^{m+},\phi^{m},t$}
\For{$ t_d \leftarrow 1,2, \cdots$} 
\State $\mathcal{L}_{ce}^{m}$ $\leftarrow$ Eq. (\ref{eq16}), and back propagation
\EndFor
\State Send $\phi _{t+1}^{m} + \mathcal{T}(t+1)$ to Sever $S$
\EndProcedure
}
\end{algorithmic}
\caption{\app{Algorithm of \ours{}.}}
\label{alg:fedgdp:main}
\end{algorithm}
}

\section{Experiments} \label{sec_exper}
We evaluate the effectiveness of our FL method on neuro-disorder classification of two real brain disease datasets: Autism brain imaging data exchange (ABIDE)~\cite{di2014autism} and Alzheimer's disease neuroimaging initiative (ADNI)~\cite{timmurphy.org}. 

\subsection{Experimental setup} \label{sec_setup}

\begin{table}[t]
\centering 
\app{
\scalebox{0.99}{
\begin{tabular}{ccccc}
\hline
Data& \multicolumn{2}{c}{ABIDE}  & \multicolumn{2}{c}{ADNI}\\ \hline
Information & ASD   & HC  & MCI& AD   \\ \hline
Subject \#& 485 & 544 & 492  & 375  \\ \hline
\begin{tabular}[c]{@{}c@{}}Gender*\\ (Female/Male)\end{tabular} & 71/414 & 145/399 & 223/269 & 166/209 \\ \hline
\begin{tabular}[c]{@{}c@{}}Age*\\ (Mean±Std)\end{tabular}& \begin{tabular}[c]{@{}c@{}}17.19\\ (±10.07)\end{tabular} & \begin{tabular}[c]{@{}c@{}}16.52\\ (±8.91)\end{tabular} & \begin{tabular}[c]{@{}c@{}}72.89\\ (±7.53)\end{tabular} & \begin{tabular}[c]{@{}c@{}}75.60\\ (±7.81)\end{tabular}\\ \hline
\end{tabular}}
\caption{\app{Demographic information of studied subjects in ABIDE and ADNI. (Std: standard deviation). ASD: autism spectrum disorder, HC: healthy control, MCI: mild cognitive impairment, and AD: Alzheimer's disease. The term $^*$ denotes that there is no significant difference ($p>0.05$) between ASD and NC, MCI and AD groups in terms of gender/age via two-tailed two sample $t$-test.} }
\label{tabledem}
}
\end{table}

\subsubsection{Datasets} \label{sec_datasets}
The demographic information of the subjects in the used datasets and are listed in Table~\ref{tabledem}.

\textbf{ABIDE} includes $1029$ subjects  from ABIDE-I and ABIDE-II, \ie $485$ ASD patients and $544$ \revision{healthy controls}\revisionfoot{\footnote{Reviewer 1, Q6}} (HC) with functional magnetic resonance imaging (fMRI) data.  
\app{The fMRI data are pre-processed using the data processing assistant for resting-state fMRI (DPARSF \cite{DPARSF})\footnote{Downloaded from http://rfmri.org/dpabi.}.} The registered fMRI volumes are partitioned into 122 regions-of-interest (ROIs) using the Bootstrap Analysis of Stable Clusters (BASC-122) template. 
We construct a $122 \times 122$ FC network for each subject, where each node is an ROI and the edge weight is the Pearson's correlation between the time series of BOLD signals of paired ROIs. 
We use the upper triangle of the fully connected matrix to represent a subject, yielding a $7,503$-dimensional feature vector.

\textbf{ADNI} 
includes $911$ subjects with T1 MRIs from ADNI-1, ADNI-GO and ADNI 2 
(\ie $375$ AD patients and $536$ mild cognitive impairment (MCI) subjects.)
\app{The structural MRI data are pre-processed  by the following operations: (1) anterior commissure-posterior commissure (AC-PC) correction, the resampled images adopt the standard 256×256×256 mode, and the N3 algorithm \cite{sled1998nonparametric}  applied to correct the non-uniform tissue intensity; (2) skull removal and cerebellectomy; (3) spatial normalization to the MNI template with 3×3×3 $mm^3$ resolution; (4) spatial smoothing using a full width at half maximum Gaussian smoothing kernel with a size of 6 mm.}
We use FreeSurfer \cite{fischl2012freesurfer} to preprocess brain tissue (white and gray matter and cerebrospinal fluid) of a T1 MRI~\cite{baronzio1999proinflammatory} and extract anatomical statistics, getting a 345-dimensional feature vector.
We also apply z-score normalization of each feature vector separately, considering the heterogeneity for different measurements.

\begin{table}[h]
\centering
\app{
\begin{tabular}{c|c}
\toprule
\textbf{Layer}              & \textbf{Details}    \\ \midrule  
1 & G-conv($D$, 256) + ELU                     \\ \hline
2 & G-conv(256, 64) + ELU                     \\ \hline
3 &  FC(64, 1)  +Sigmoid                     \\ \hline
4  & Random-vector(4)  \\ \hline
5 & Linear(68, 128) + ReLU+BN(128)                   \\ \hline
6 & Linear(128, 256)  + ReLU+BN(256)                   \\ \hline
7 & FC(256, $D$) +   tanh                  \\ \hline
8 & Linear($D$, 32)  ReLU                   \\ \hline     
9 & FC(32, 2)                 \\ \bottomrule 
\end{tabular}
}
\caption{\app{Network architecture of the missing node generator.}}
\label{tab_arch1}
\end{table}

\begin{table}[h]
\centering
\app{
\begin{tabular}{c|c}
\toprule
\textbf{Layer}              & \textbf{Details}    \\ \midrule  
1  & SN-Linear($D$, 128) + ReLU \\ \hline
2 & SN-Linear(128, 32) + ReLU                     \\ \hline
3 & SN-Linear(32, 1)         \\ \bottomrule
\end{tabular}
}
\caption{ \app{Network architecture of the discriminator.}}
\label{tab_arch2}
\end{table}

\begin{table}[h]
\centering
\app{
\begin{tabular}{c|c}
\toprule
\textbf{Layer}              & \textbf{Details}    \\ \midrule  
1  & G-conv($D$, 64) + ELU \\ \hline
2 & G-conv(64, 32)                 \\ \hline
3 & FC(32, 2)         \\ \bottomrule
\end{tabular}
}
\caption{\app{Network architecture of  the GCN  node  prediction  module.}}
\label{tab_arch3}
\end{table}

\begin{table*}[!ht]
\small
\vspace{-1mm}
\centering
\begin{tabular}{ccccccccccc}
\toprule
&\multicolumn{5}{c}{\textbf{ABIDE}}& \multicolumn{5}{c}{\textbf{ADNI}} \\
\cmidrule(r){2-6} \cmidrule(r){7-11} Method&Accuracy&AUC&Precision&Recall&F1-score&Accuracy&AUC&Precision&Recall&F1-score\\
\midrule

 ~~\multirow{2}{*}{LocalGCN} &0.600  & 0.598  & 0.580  & 0.559  & 0.560   &0.703  & 0.695  & 0.666  & 0.637  & 0.643   \\ 
&($\pm$0.012) & ($\pm$0.011) & ($\pm$0.012) & ($\pm$0.017) & ($\pm$0.012)&($\pm$0.016) & ($\pm$0.018) & ($\pm$0.022) & ($\pm$0.037) & ($\pm$0.028)\\
\hline
 ~~\multirow{2}{*}{CentralGCN} &0.655  & 0.654  & 0.635  & 0.633  & 0.633   &0.757  & 0.750  & 0.734  & 0.691  & 0.710    \\ 
&($\pm$0.008) & ($\pm$0.008) & ($\pm$0.010) & ($\pm$0.015) & ($\pm$0.011)&($\pm$0.007) & ($\pm$0.006) & ($\pm$0.013) & ($\pm$0.014) & ($\pm$0.008)\\
\hline
\midrule
 ~~\multirow{2}{*}{FedMLP}   &0.630  & 0.627  & 0.618  & 0.573  & 0.591   &0.731  & 0.721  & 0.625  & 0.631  & 0.625       \\ 
&($\pm$0.011) & ($\pm$0.011) & ($\pm$0.017) & ($\pm$0.029) & ($\pm$0.015)&($\pm$0.026) & ($\pm$0.030) & ($\pm$0.081) & ($\pm$0.077) & ($\pm$0.079)\\
\hline
 ~~\multirow{2}{*}{FedGCN}  &0.644  & 0.643  & 0.625  & 0.616  & 0.613  &0.742  & 0.735  & 0.704  & 0.693  & 0.692  \\ 
&($\pm$0.009) & ($\pm$0.009) & ($\pm$0.011) & ($\pm$0.022) & ($\pm$0.012)&($\pm$0.007) & ($\pm$0.009) & ($\pm$0.010) & ($\pm$0.014) & ($\pm$0.009)\\ 
\hline
 ~~\multirow{2}{*}{FedSage+}&0.658  & 0.655  & 0.641  & 0.627  & 0.626   &0.751  & 0.744  & 0.722  & 0.699  & 0.700   \\ 
&($\pm$0.006) & ($\pm$0.008) & ($\pm$0.013) & ($\pm$0.017) & ($\pm$0.009) & ($\pm$0.008) & ($\pm$0.010) & ($\pm$0.009) & ($\pm$0.015) & ($\pm$0.010)\\
\hline
 ~~\multirow{2}{*}{\ours (ours)}&\textbf{0.667}  & \textbf{0.663} & \textbf{0.647}  & \textbf{0.640}  & \textbf{0.637}  &\textbf{0.758}  & \textbf{0.754}  & \textbf{0.725}  & \textbf{0.721}  & \textbf{0.717}      \\ 
&($\pm$0.006)&($\pm$0.007)&($\pm$0.009)&($\pm$0.014)&($\pm$0.007)&($\pm$0.007) & ($\pm$0.007) & ($\pm$0.008) & ($\pm$0.017) & ($\pm$0.011)\\
\bottomrule
\end{tabular}
\caption{Comparison of performance (\ie Accuracy, AUC, Precision, Recall and F1-score) in the format of mean and standardard deviation on both ABIDE and ADNI datasets.  CentralGCN indicates pulling data together.  The best performance among FL settings is highlighted.}
\vspace{-1mm}
\label{tab51}
\end{table*}

\subsubsection{Model training Setting-up} We list the detailed settings for FL and implementation platforms as follows.

\noindent  \textbf{Federated learning settings.}
Given the population size of ABIDE and ADNI, we set the number of institutions to 5 (\ie $M = 5$) for both datasets for all methods except CentralGCN which is trained on the whole dataset. 
For both datasets, we equally and randomly divide subjects into 5 subsets and assign each subset to an institution\footnote{The main goal of our work is not to handle the significant distribution heterogeneity of data collected in different institutions. To control this factor, we randomly divide the samples instead of clustering them by institution.}. Specifically, each institution has 206 subjects for the ABIDE dataset and 182 subjects for the ADNI dataset. Following Kipf \etal\cite{kipf2016semi}, we use full-batch training on our small local graphs. We use Adam as our optimizer and set the learning rate as 0.001 for all the experiments. We detail the training iterations for different methods in Sec.~\ref{sec_comp}. 

%
\noindent  \textbf{Implementation platform and experimental factors control.} All experiments are implemented in PyTorch and conducted on a server with 8 NVIDIA GeForce 3090 GPUs (24 GB memory for each GPU). 
\revision{The labeled rate is set to 80\% and the training/testing data is split according 5-fold cross-validation.}\revisionfoot{\footnote{Reviewer 3, Q7}}
We obtain the author-verified codes for all comparison methods and follow the advice of parameter settings in the corresponding literature so that all comparison methods achieve the best performance on each dataset. Additionally, all methods (including our method) use the same settings for the original graph structure, the train/test partitioning, the dimension of the networks and the training procedures. 
\app{Besides, Tables \ref{tab_arch1}-\ref{tab_arch3} present the used network architectures of the missing node generator, the discriminator and the GCN node prediction module, which are implemented with the PyTorch framework. 
A graph convolution layer is represented by ‘G-conv’, and a batch normalization layer is denoted by ‘BN’. 
We use ‘Linear’ to denote a linear transformation layer
and ‘FC’ to denote a fully-connected layer for classification.
The ‘ReLU’ is used as the non-linearity function, and ‘ELU’ denotes the exponential linear unit.
The sigmoid function is represented by ‘sigmoid’, and the hyperbolic tangent function is denoted as  ‘tanh’.
In addition, we show the channel dimension in the bracket where $D$ denotes the input dimension.
}

\subsubsection{Performance evaluation}
The diagnosis results of all methods are evaluated by five evaluation metrics, including \textit{Accuracy, Area under the ROC Curve (AUC), Precision, Recall and F1-score}. For all of these metrics, a higher value means better performance. In all experiments, we performed 5-fold cross-validation and repeated the experiments 5 times for each method with random seeds to report their average performances and corresponding standard deviation. We conduct significance testing using the two sample t-test.

\subsection{Comparison with alternative methods}
\label{sec_comp}
\subsubsection{Alternative methods}


We compare with two centralized methods and three FL methods. 
The details of the comparison methods are listed as follows:

\textbf{CentralGCN}\cite{parisot2018disease} assumes data are centralized and trains a GCN model on the original global graph. The graph construction strategy is the same as ours.

\textbf{LocalGCN} follows the model architecture in CentralGCN, but trains the GCN model exclusively on each local graph under non-FL setting. 

\textbf{FedMLP}\cite{li2020multi}  applies the FedAvg  strategy \cite{mcmahan2017communication} to train the basic MLP model with individual flattened brain connectivity matrix, without considering the population graph information.

\textbf{FedGCN} is built on LocalGCN by simply employing FedAvg \cite{mcmahan2017communication} on local GCN.

\textbf{FedSage+}\cite{zhang2021subgraph}  trains a missing node generator and a GraphSage classifier for each local graph to conduct FL. In particular, the missing node generator is a regression model only and it is designed to generate the features of missing nodes without considering generating the phenotype information and augmenting edges.

\begin{figure*}
\scalebox{0.85}{
\hspace{-1mm}\subfloat[ABIDE]{\scalebox{0.3}{\includegraphics{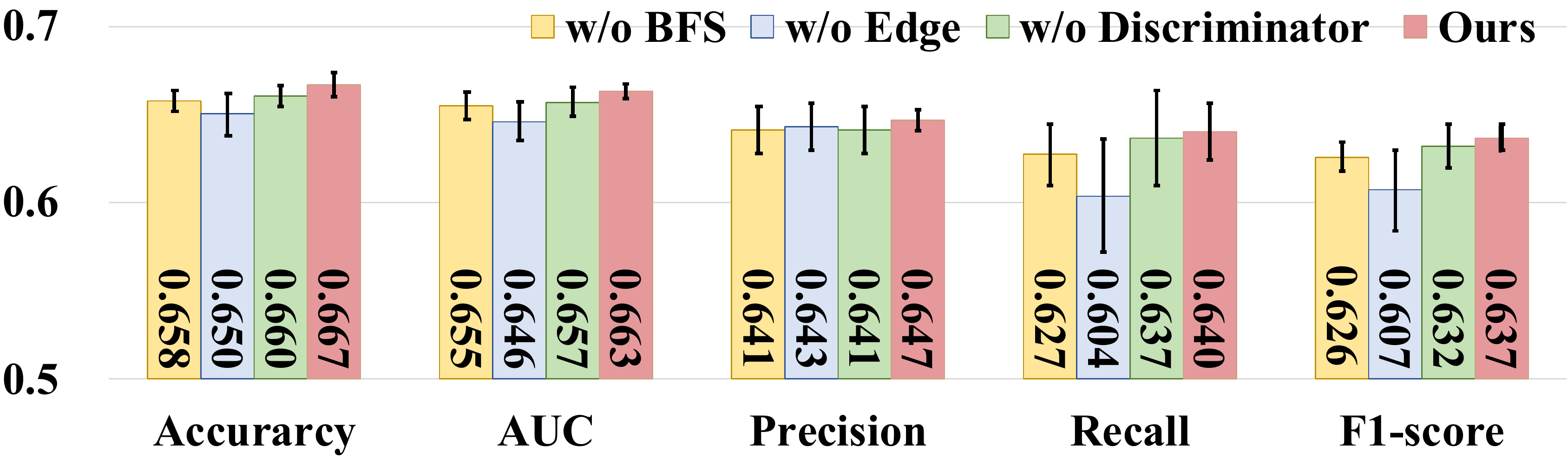}}}\subfloat[ADNI]{\scalebox{0.3}{\includegraphics{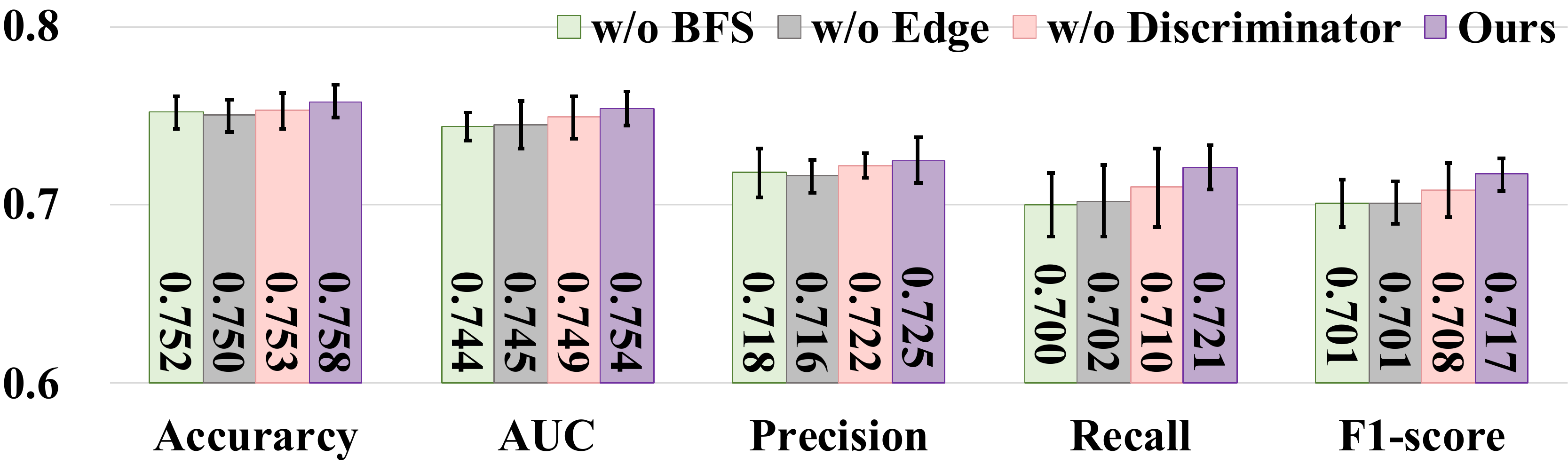}}}}
\caption{Ablation analysis of our method on two datasets.}
\label{tab52}
\end{figure*}





\begin{figure*}
\scalebox{0.85}{
\hspace{-1mm}\subfloat[ABIDE]{\scalebox{0.3}{\includegraphics{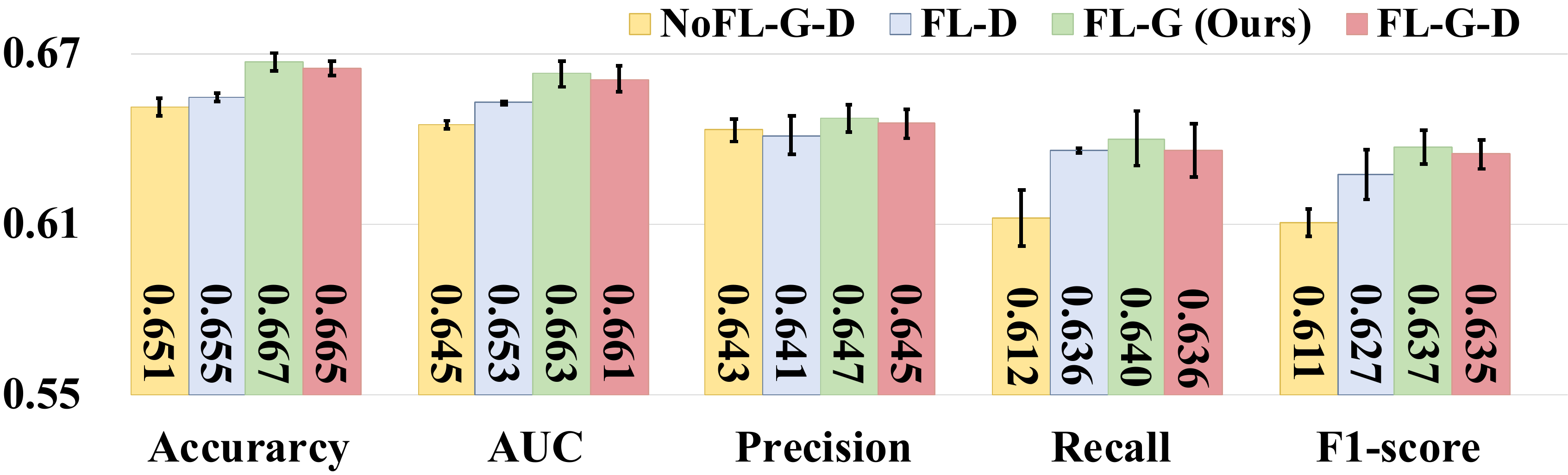}}}\subfloat[ADNI]{\scalebox{0.3}{\includegraphics{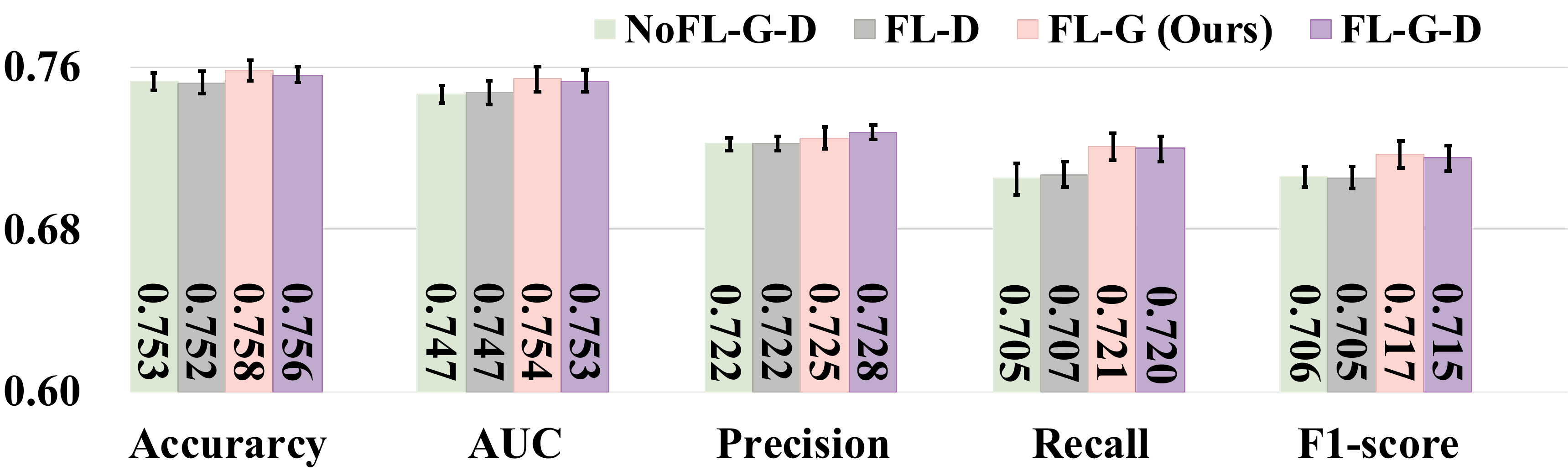}}}}
\caption{Results of our FedNI using different network inpainting module training strategies on two datasets.}
\label{fig52}
\end{figure*}

The learning epochs for centralized training methods (CentralGCN and LocalGCN) are set to 100. For FL methods, we need to set both the local update epochs between each global aggregation $E$ and the local-global model communication round $T$. For FedMLP, FedGCN, and FedSage+, we set $E=10$ and $T = 10$. As our method \ours~uses a two-phase training strategy, we set $E=10$ and $T = 30$ for federated network inpainting, then $E=10$ and $T = 10$ for federated GCN node classification. We choose a two-phase strategy as it can stabilize GAN training by reducing the complexity of optimization. Also, we have observed better and stable performance compared to the end-to-end training strategy.

\begin{figure*}
\scalebox{0.85}{
\hspace{-1mm}\subfloat[ABIDE]{\scalebox{0.3}{\includegraphics{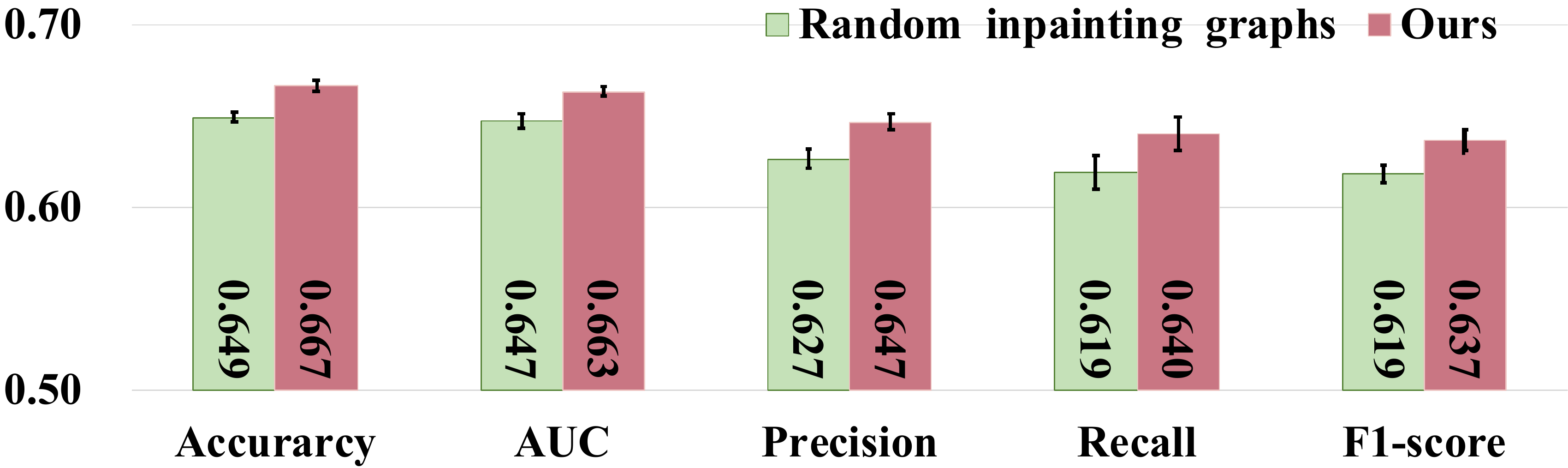}}}\subfloat[ADNI]{\scalebox{0.3}{\includegraphics{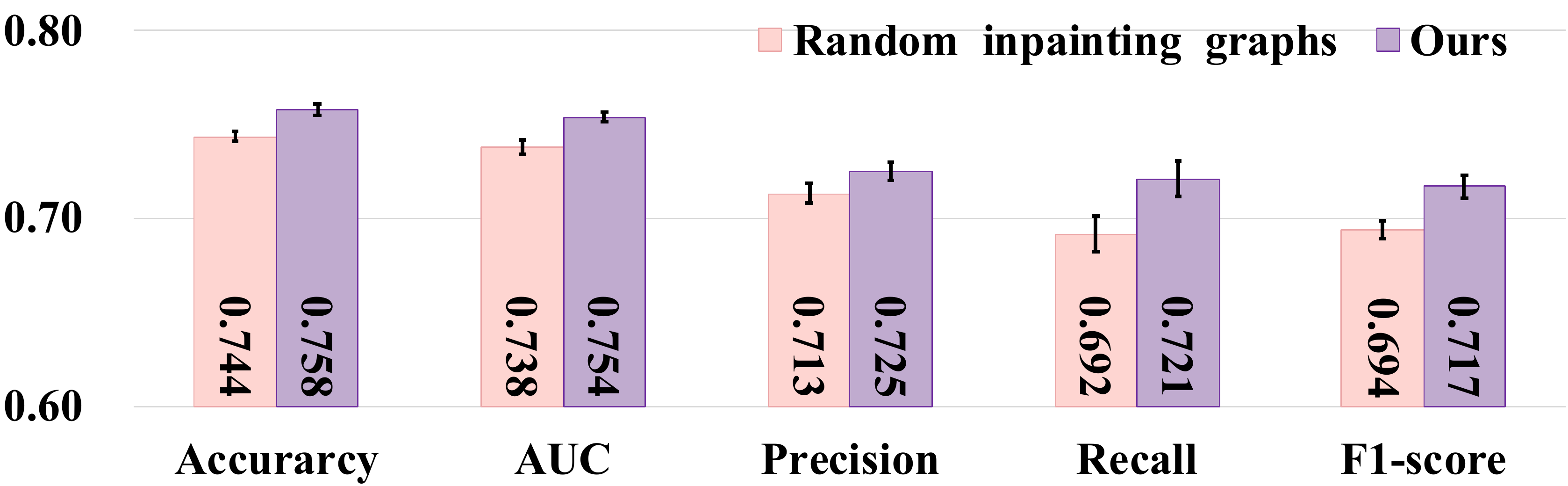}}}}
\caption{Comparison with random inpainting graphs on two datasets.}
\label{fig55}
\end{figure*}

\subsubsection{Results and analysis} \label{sec_IV-B}

Table \ref{tab51} summarizes the classification performance over institutions of all methods on the two neurological disease datasets.
First, our method outperforms all methods \revision{in terms of accuracy, AUC, precision,} recall and F1-score on two datasets.
\revision{Meanwhile, we find that the improvements are significant (with $p<0.05$ via t-test) compared with comparison methods.
Particularly, first, our method, \ours, improves by $9.2\%$ ($p=8.53e^{-5}$) on average, compared to the vanilla population-based disease classification on siloed institutions (\ie LocalGCN).} This indicates the necessity of deploying FL to encourage multi-institutional collaborative learning with privacy constraints. \ours's performance may even be slightly better than that of CentralGCN, owing to the unique FL optimization scheme that alleviates the effect of noise data on model updating direction. 
\revision{it is also worth noting that the sample of neuroimaging data is limited, thus the neighbors generated by FedNI not only inpaint the distribution differences between local clients, but also involve more training nodes at the global range to get better representation.}\revisionfoot{\footnote{Reviewer 2, Q3}}
\revision{Second, \ours{} outperforms FedGCN by 3.61\% ($p=1.27e^{-3}$) on ABIDE dataset and 3.07\% ($p=2.23e^{-3}$) on ADNI dataset.
To this end, \ours{} also achieves better performance than the best FL competitor FedSage+. Specifically, compared to FedSage+, our method achieves an average  improvement of 1.2\% ($p=0.009$), 1.4\% ($p=0.013$), 0.6\% ($p=0.073$), 2.6\% ($p=0.003$), and 2.1\% ($p=0.005$) in terms of accuracy, AUC, precision, recall and F1-score, respectively.}\revisionfoot{\footnote{AE, Q1 and Reviewer 2, Q3}}
These results suggest that the effectiveness of our proposed \ours{} federated network inpainting design can achieve better federated graph learning performance on small siloed population graphs than the state-of-the-art (\ie FedSage+).

\subsection{Ablation study}
\label{sec_ablation}

In this section, we conduct extensive ablation studies to demonstrate the necessity of the three essential techniques in the network inpainting module (see Sec.~\ref{sec_ablation_components}); verify the rationality of federatively training generators only (see Sec.~\ref{sec_diff_train}); and demonstrate the power of complete, more reliable local graphs through the proposed learning mechanism (see Sec.~\ref{sec_random_inpainting}). 

\subsubsection{Ablation on BFS, edge prediction, and discriminator}
\label{sec_ablation_components}
The proposed FL local network inpainting module consists of three essential components, including applying BFS to generate an incomplete graph, incorporating phenotype data for edge prediction, and using SN-GAN discriminator to improve feature fidelity. To substantiate the superiority of these components, comprehensive ablation studies are performed by removing each component from our method and Fig. \ref{tab52} shows how each of the components contributes to the final performance.

\noindent  \textit{\textbf{Effectiveness of BFS.}} 
To evaluate the effectiveness of BFS algorithm (\ie Eq. (\ref{eq3})), we report the performance of hiding nodes by the BFS algorithm and the random method, respectively, in  Fig. \ref{tab52}.
It can be seen that when using BFS algorithm for hiding nodes, the average performance of AUC gains 1.0\% and 1.1\% comparing with random mask
on the ABIDE dataset and the ADNI dataset, respectively.


\noindent   \textit{\textbf{Effectiveness of edge prediction.}} 
Edge information is an essential element for GCN model. A missing edge and its feature can result in incomplete ego networks on the graph, thus hurting performance of the disease prediction. In this part, we investigate the effectiveness of  utilizing phenotype data to predict edges. To achieve this, we replace the edge reconstruction operation (\ie Sec. \ref{sec_phe}) in our missing node generator method with adding binarized links between the ego nodes and their missing neighbors~\cite{zhang2021subgraph}. 
As can be seen from Fig.~\ref{tab52}, our method can achieve better performance on all evaluation metrics. For example, we obtain the AUC result of 0.663, which is higher than that (\ie 0.646) of the ablation without generating new edges.
It can be concluded that generating edges based on phenotype predictions is essential for generating a high-quality graph.

\begin{figure}[!b]
\vspace{-4mm}
\centering
\hspace{-6mm}{\scalebox{0.275}{\includegraphics{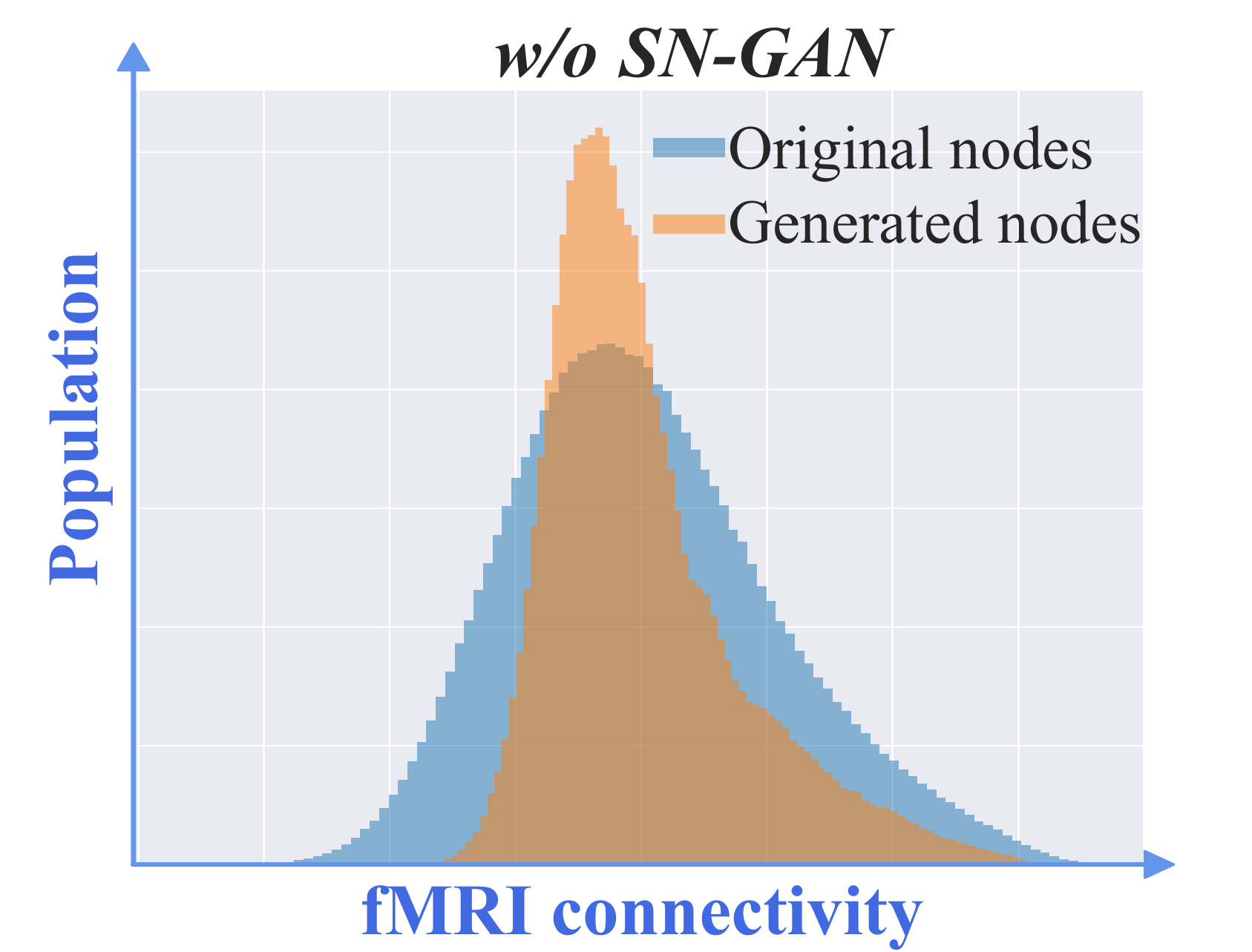}}}{\scalebox{0.275}{\includegraphics{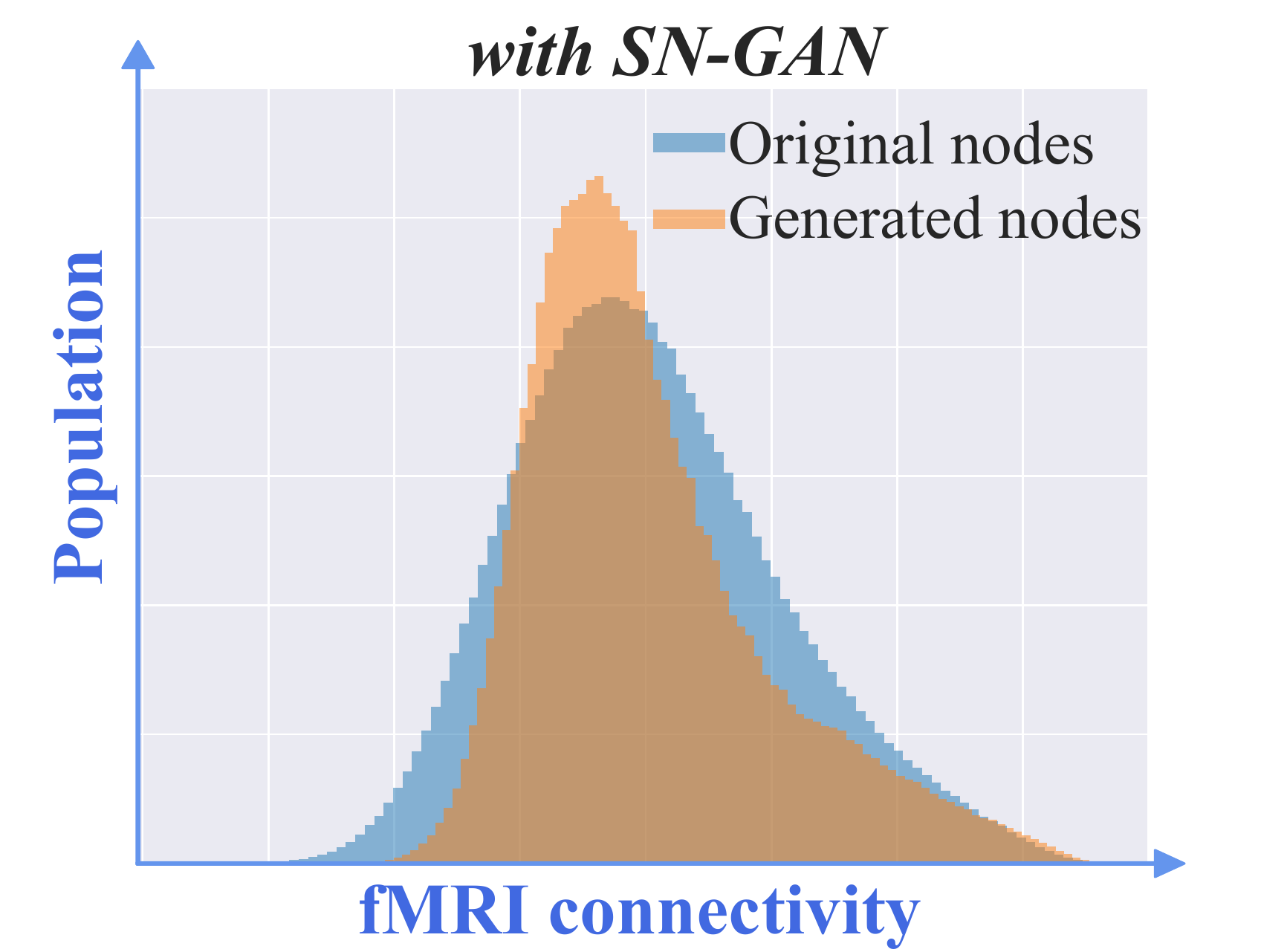}}}
\vspace{-2mm}	
\caption{Feature distribution of original nodes and generated nodes in two cases: (i) Without SN-GAN. (2) With SN-GAN.}
		\label{figgan}
\end{figure}
\begin{figure}
\vspace{-4mm}
\subfloat[ABIDE]{\scalebox{0.2}{\includegraphics{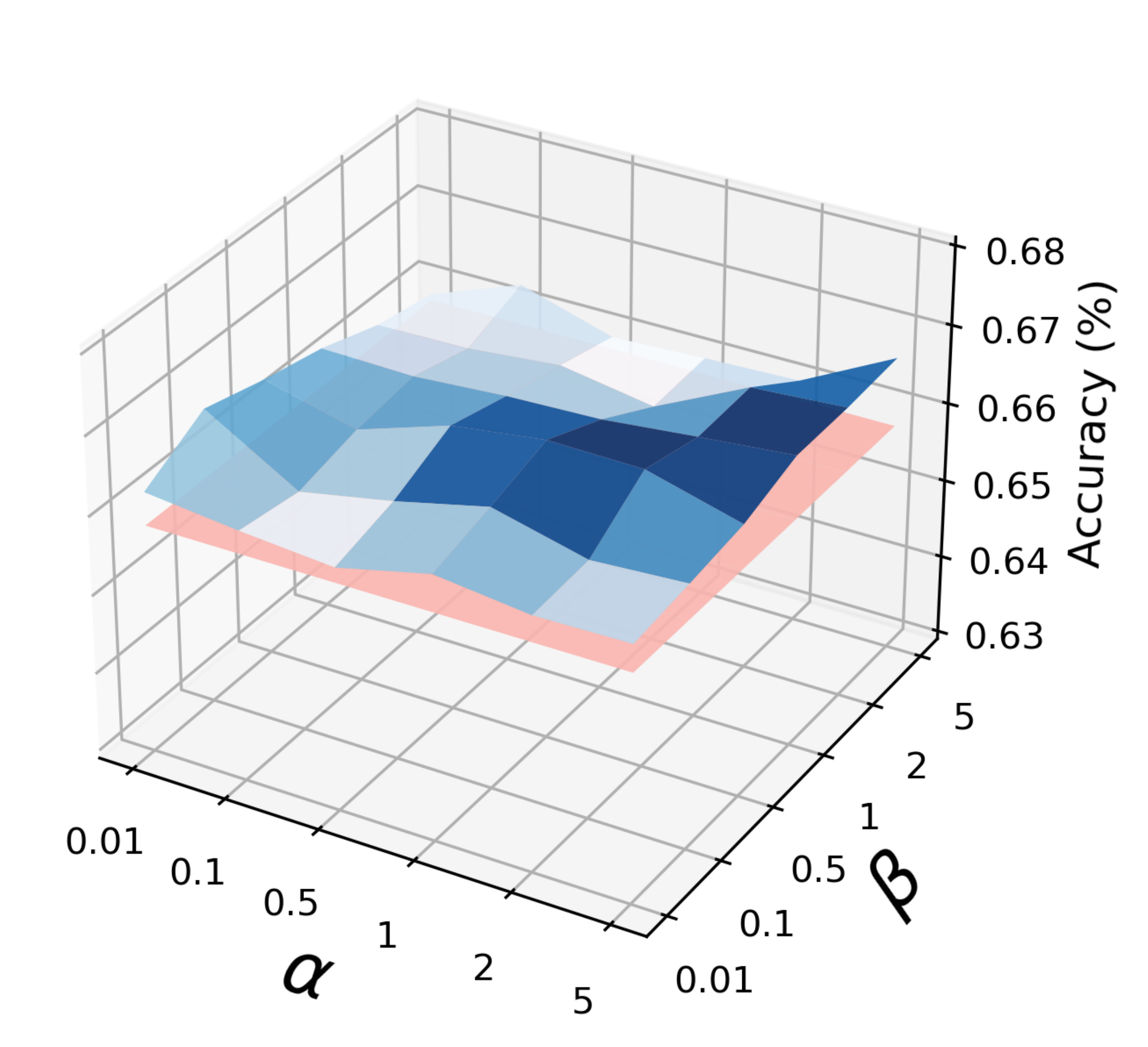}}}
\subfloat[ADNI]{\scalebox{0.2}{\includegraphics{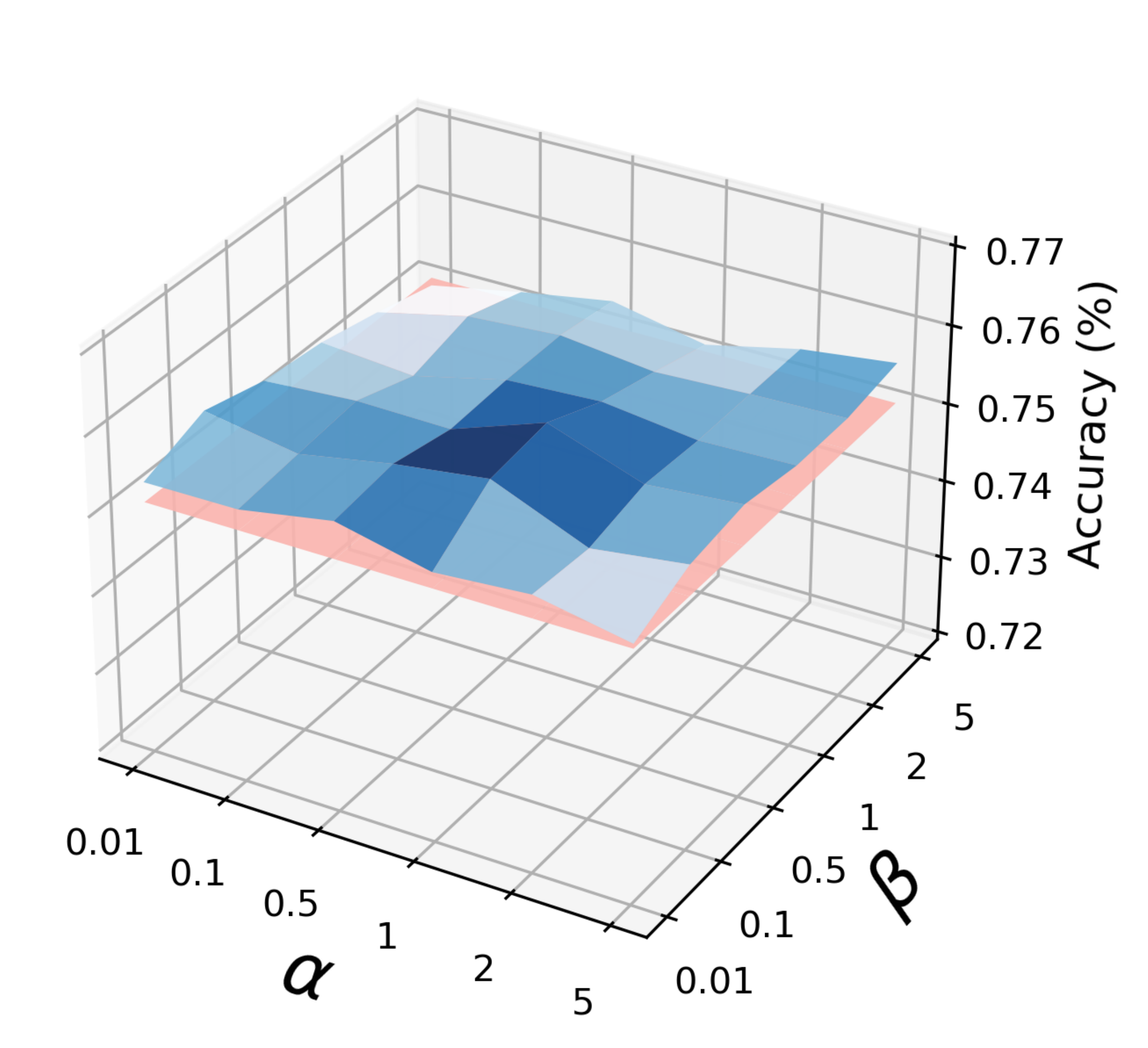}}}
	
\caption{Averaged testing accuracy over all the institutions of our method (blue surface) with different parameter settings
(\ie $\alpha$ and $\beta$). The red plane represents the best results of Fedsage+. The blue surface is always on top of the red plane.}
\label{fig_parameter}
\end{figure}

\noindent  \textit{\textbf{Effectiveness of discriminator.}} 
We perform federated graph inpainting without the discriminator (\ie Eq.~(\ref{eq11}) and Eq.~(\ref{eq13})) to further verify its effectiveness in our method. 
From Fig. \ref{figgan}, we can intuitively see the feature distribution of original nodes and generated nodes in two cases. 
This figure suggests that our method with discriminator can significantly generate a more stable feature distribution, compared with the method without discriminator. 
The main reason could be that the discriminator use spectral normalization to make discriminator satisfy the Lipschitz hypothesis, thereby making the discriminator model more stable. 
Furthermore, through this rough comparison, it can be seen that our \revision{FedNI gets the best accuracy and precision} results in  Fig~\ref{tab52}. 
This suggests that our method with discriminator can significantly generate more stable and realistic nodes, compared to the method without discriminator. 

\subsubsection{Different network inpainting module training strategies} \label{sec_inpain_stratege}
To further investigate the effectiveness of applying FL on the network inpainting module, we conduct experiments with four variants of training strategy on our network inpainting module: 
(i) Training the generator model (\ie G) and the discriminator model (\ie D) of network inpainting module under non-FL setting (\ie NoFL-D-G); (ii) Only the discriminator model under FL setting (\ie FL-D); (iii) Only the generator model under FL setting (\ie FL-G); (iv) Training the generator model and the discriminator model under FL setting (\ie FL-D-G).
Fig.~\ref{fig52} shows the results of the proposed network inpainting module with four variants of FL training strategy, from which we observe the following: 
(1) Applying FL on generator model (\ie FL-G) generally outperforms other variants of training strategy (\eg NoFL-D-G and FL-D).  For example, the FL-G method achieves significantly better performance (with $p<0.05$ via t-test) compared with NoFL-D-G and FL-D \revision{in terms of accuracy, AUC, precision,}\revisionfoot{\footnote{Reviewer 1, Q5}} recall and F1-score. 
The main reason could be that applying FL on the generator model can lead the local generator model to output node features that are more similar with the node features of other organizations.
(2) There was no significant improvement when using FL on the discriminator, as seen from comparing FL-G with FL-G-D and comparing NoFL-D-G with FL-D. The reason might be that the generator can directly influence the generated features, while the discriminator influences it indirectly.
\revision{The intuition is that the generator should predict missing nodes following the data distribution of the global population network, while local discriminators can fit the heterogeneity of local graphs better.}\revisionfoot{\footnote{Reviewer 1, Q2}}
These observations further verify the effectiveness of our strategy (FL-G only) for network inpainting.

\label{sec_diff_train}

\subsubsection{Comparison with random inpainting}
\label{sec_random_inpainting}
In practice, the generated missing nodes and edges might be inaccurate.
To further illustrate the strength of the proposed network inpainting method, we conduct experiments on network inpainting by our method and random inpainting method.
Specifically, for random network inpainting method, we uniformly generate the same number of missing nodes as our method, and randomly generate node features based on the original feature distribution (\ie normal distribution) as well as edges.
Fig.~\ref{fig55} shows the results of network inpainting using the random network inpainting method and our method in ABIDE and ADNI datasets. It can be seen from Fig.~\ref{fig55} that our FedNI achieves consistently better performance compared to the random inpainting method. 
The experiments show that the missing nodes generated by our proposed network inpainting method are accurate and provide more useful information for the local GCN classifier to improve the performance.

\revision{ 
\subsubsection{The scalability of FedNI} \label{sec_scalability}
We conduct an experiment to investigate the scalability of \ours{} in terms of the number of FL clients (increasing the number of FL clients and keeping the number of nodes in local clients fixed).
In this way, we fix the number of nodes in each local client (\ie 10\% of nodes in the dataset), and increase the numbers of the clients, \ie from 2 to 10.  
The more clients, the more training samples are available for global model learning their patterns. As shown in Fig. \ref{fig_numberFL2_main},  we observe that involving more clients can improve performance and maintain consistency.
On the basis of these results, we can concluded that \ours{} has a promising scalability, as increasing the number of clients in global scope, resulting a better performance. In this case, if we want obtain a better result for practical application, we can achieve this goal by involving more clients.
Importantly, we report the computation cost of our result in Table. \ref{tab_cost_main}. Intuitively, this makes sense for the scalability, as low computation cost allows involving more FL clients to get better results.
\begin{figure}[!t]
\centering
{
\subfloat{\scalebox{0.35}{\includegraphics{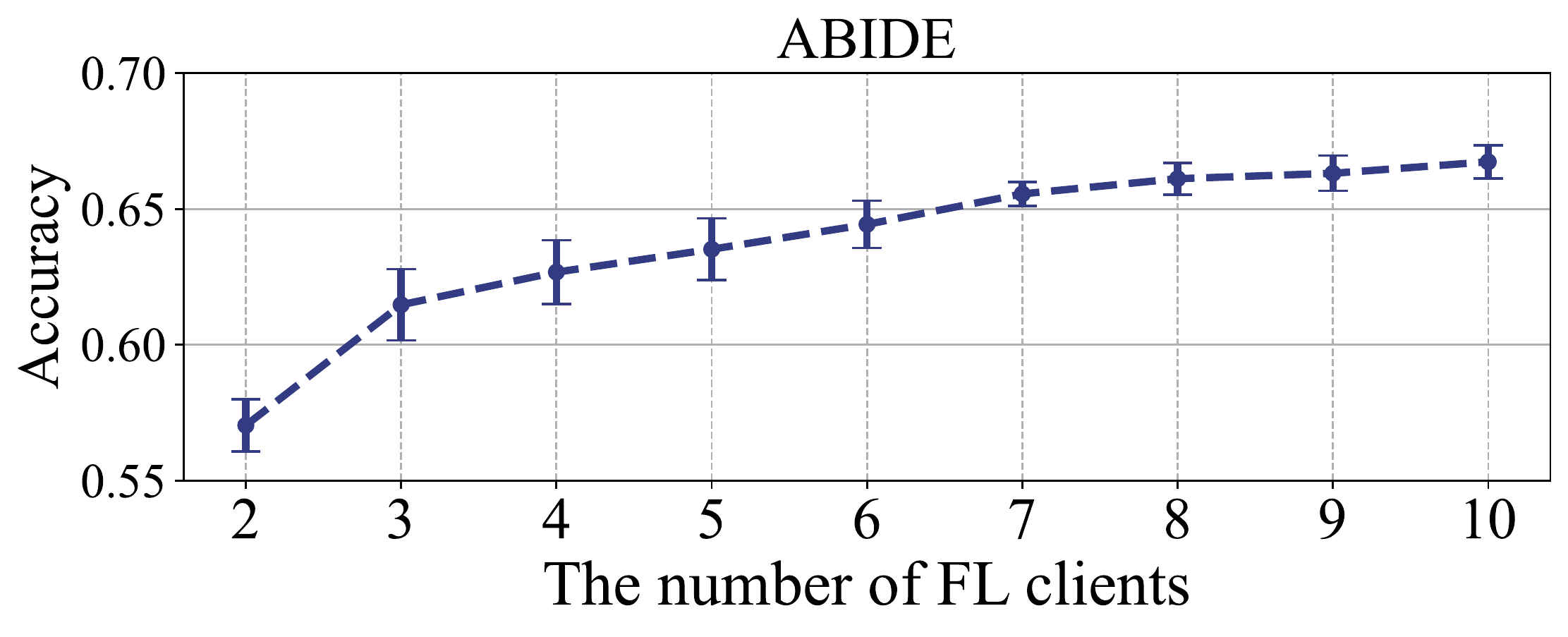}}}\\\vspace{-1mm}
\subfloat{\scalebox{0.35}{\includegraphics{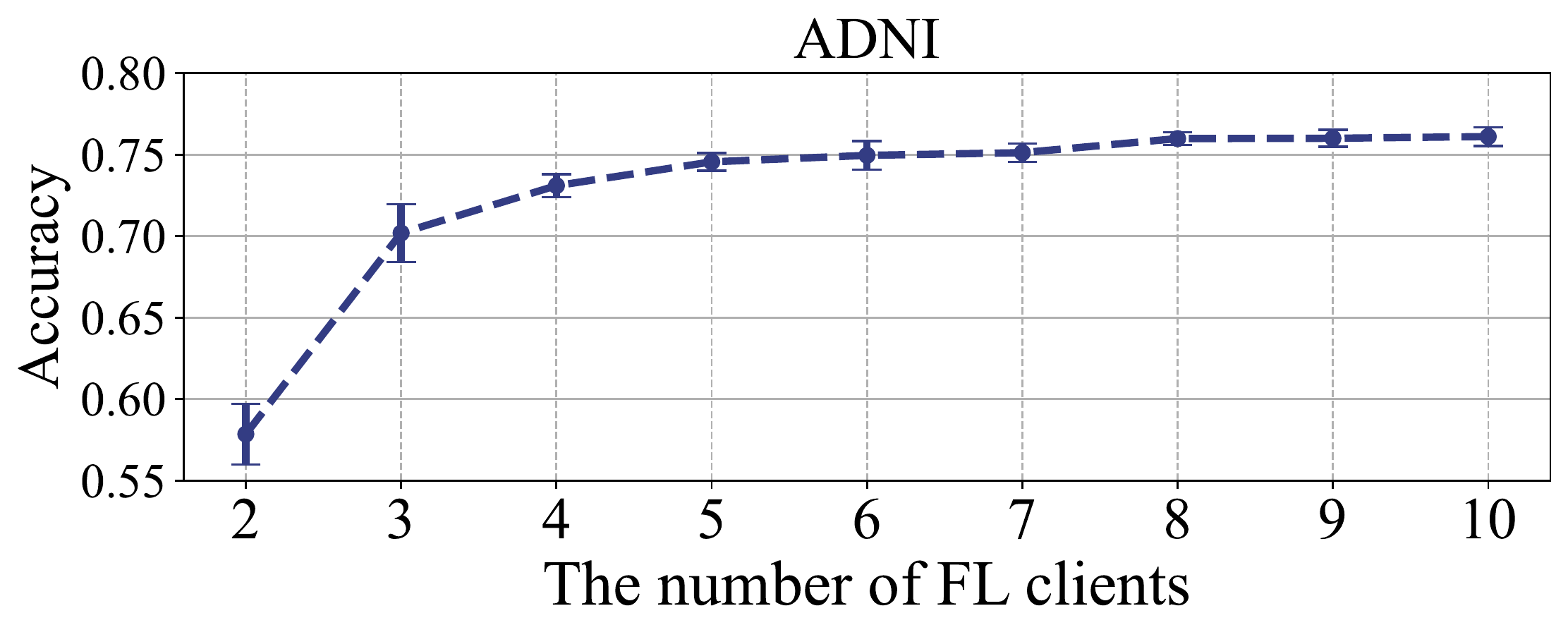}}}}
\caption{\revision{Classification accuracy of \ours{} with fixed number of nodes in local clients at different numbers of FL clients (from $2$ to $10$) on two datasets.}}
\label{fig_numberFL2_main}
\end{figure}

\subsubsection{The computational cost} \label{sec_cost}
The experiment of computational cost is implemented with PyTorch (vision 1.9) framework and reported by four metrics, including \textit{FLOPs} ($G$), \textit{Params}  ($M$), \textit{Training time} ($s$) and \textit{Total times} ($s$). Note that, the proposed \ours{} includes two FL phases (\ie \textit{Federated network inpainting} and \textit{Federated GCN  node classification}), we report the computational cost for each phases as well as the \textit{total time} (The \textit{total time} including \textit{training times} of two phases and \textit{graph merge} process as we described in Sec. \ref{sec_fgel}).
For \textit{Federated network inpainting} phase, the input of the model is one local graph for a client (\ie a graph with 200 nodes where each nodes contain a 7381 dimensional feature vector). 
For \textit{Federated GCN  node classification} phase, the input of the model is one local fused graph for a client (\ie a graph with 300 nodes where each nodes contain a 7381 dimensional feature vector). 
From Tab. \ref{tab_cost_main}, we observed that both two phases of \ours{} are low computational. Although \textit{Federated network inpainting} phase brings extra computational cost, it is acceptable since they are in the same levels of computational cost. The results indicate that \ours{} has the has the potential potential to be applied in practice and good scalability.
}\revisionfoot{\footnote{Reviewer 3, Q2}}
\begin{table}[h]
\centering
\revision{
\small
\begin{tabular}{l|cc}
\hline
 & \multicolumn{1}{c|}{\begin{tabular}[c]{@{}c@{}}Federated network \\ inpainting\end{tabular}} & \begin{tabular}[c]{@{}c@{}}Federated GCN \\ node classification\end{tabular} \\ \hline
FLOPs ($G$) & \multicolumn{1}{c|}{0.98} & 0.26 \\ \hline
Params ($M$) & \multicolumn{1}{c|}{3.1} & 0.47 \\ \hline
Training time ($s$) & \multicolumn{1}{c|}{66} & 11 \\ \hline
Total time ($s$) & \multicolumn{2}{c}{89} \\ \hline
\end{tabular}
}
\caption{\revision{The computational cost of \ours{}.}}
\label{tab_cost_main}
\end{table}

\subsection{Hyperparameter Discussion}

\subsubsection{Parameter sensitivity analysis} \label{sec_parameter}
In our experiment, we use the default non-weighted losses in Eq. (\ref{eq12}), namely $\alpha = 1$ and $\beta = 1$ for both datasets. We investigate the sensitivity of the weighted coefficient (\ie $\alpha$ and $\beta$) of our method in Eq. (\ref{eq12}). 
We vary the values of $\alpha$ and $\beta$ from $10^{-2}$ to $5$  and report the mean values of accuracy within 5 independent runs. Fig. \ref{fig_parameter} indicates that our method is insensitive to $\alpha$ and $\beta$ and consistently better than FedSage+'s best performance (\ie red plane in Fig. \ref{fig_parameter}).
This is because our model has a well-designed feature reconstruction framework and is robust.

\begin{figure}[!t]
\centering
{
\subfloat{\scalebox{0.35}{\includegraphics{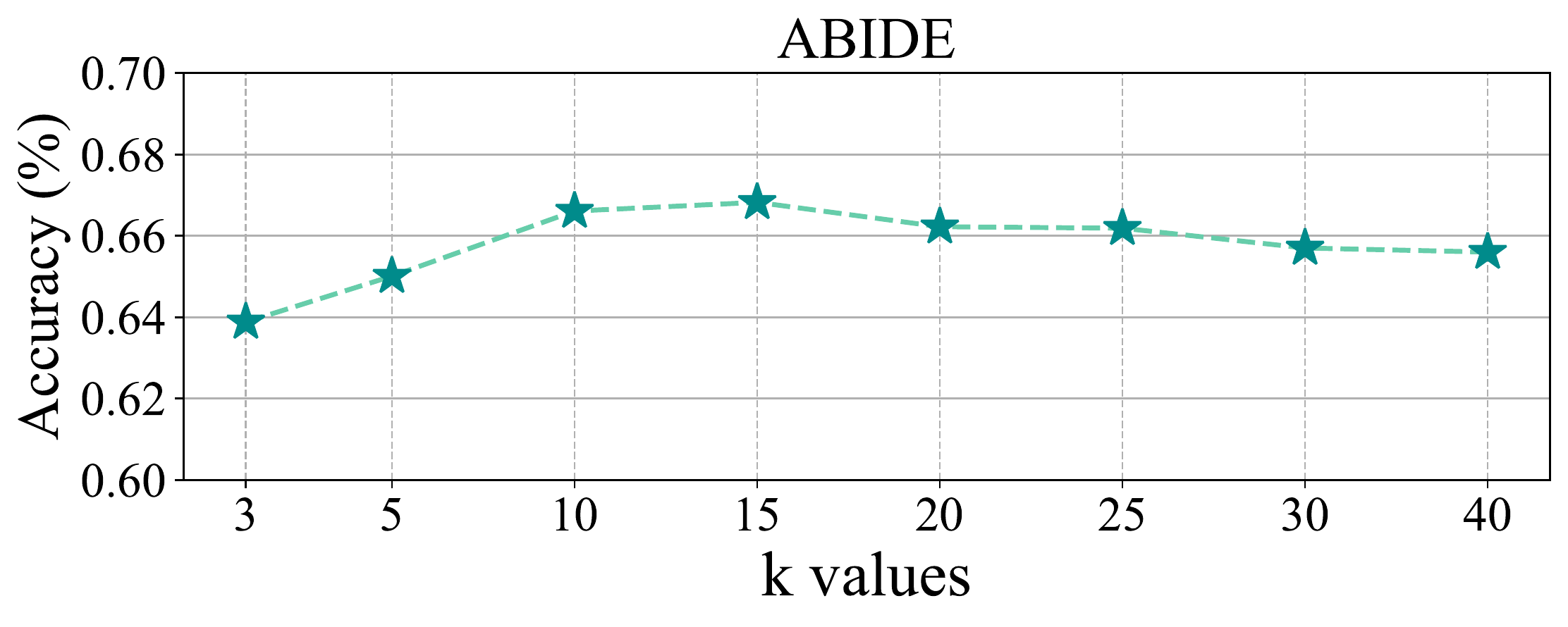}}}\\\vspace{-1mm}
\subfloat{\scalebox{0.35}{\includegraphics{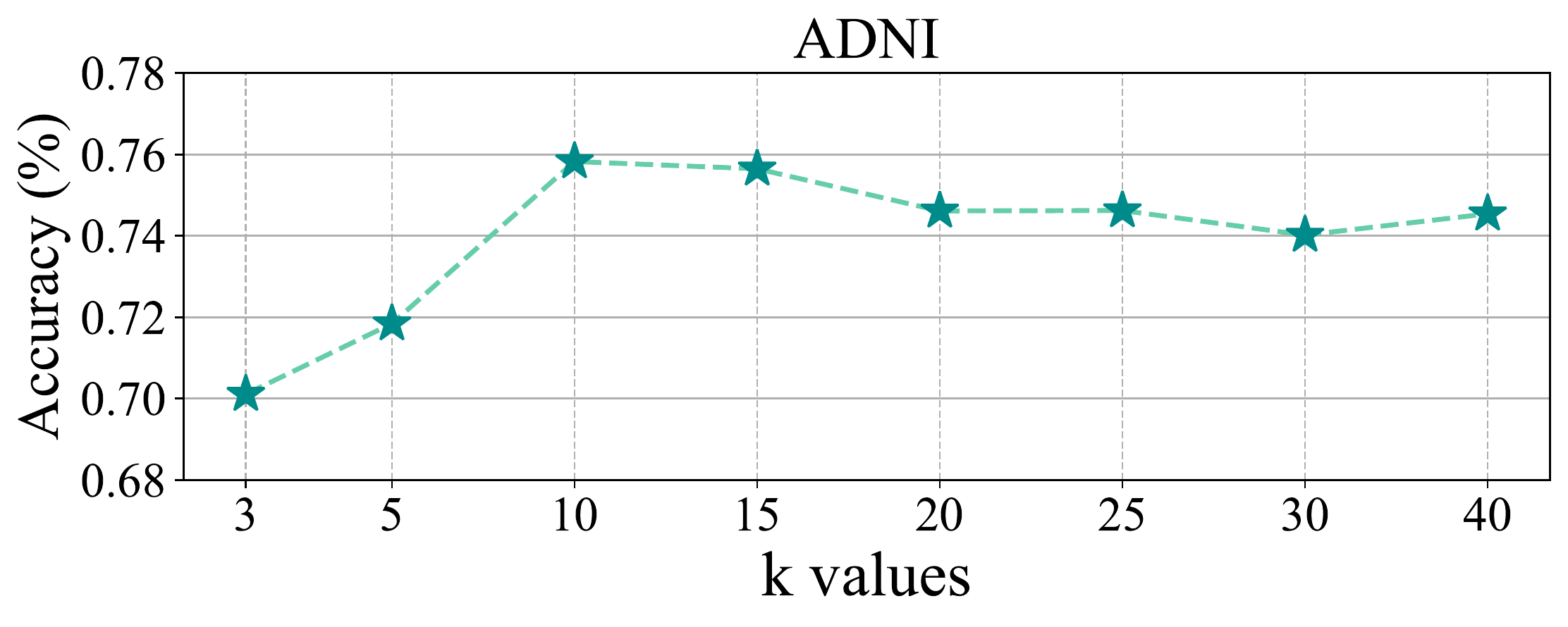}}}}
\caption{\revision{Classification results of \ours{} with different values of hyperparameter $k$.}}
\label{fig_kvalues_main}
\end{figure}

\revision{We further conduct an experiment to investigate the hyperparameter $k$. The hyperparameter $k$ in Section. \ref{sec_A} is the number of nearest neighbors. We set the range of the $k$ as $\{3, 5, 10, 15, 20, 25, 30, 40\}$, and the results of different values of hyperparameter $k$ are summarized in Fig. \ref{fig_kvalues_main}. We can observe that the classification accuracy of our  method  increased with the increasing values of $k$, \ie from $k = 3$ to $k = 10$, and increasing the value of $k$ beyond a threshold can slight hurt the performance, \ie, from $k = 15$ to $k = 40$. The reason is that a small $k$ value cannot fully exert the ability of neighbors, and a large value of $k$  may cause noise neighbors and over-smooth problem in the GCN. Moreover the optimal $k$ value in our method can be easily found (\ie, around $k = 10 $) on both two datasets.}\revisionfoot{\footnote{Reviewer 3, Q8}}

%




\subsubsection{Local updating epochs} \label{sec_epochs}
\begin{figure}[h]
\subfloat[ABIDE]{\scalebox{0.4}{\includegraphics{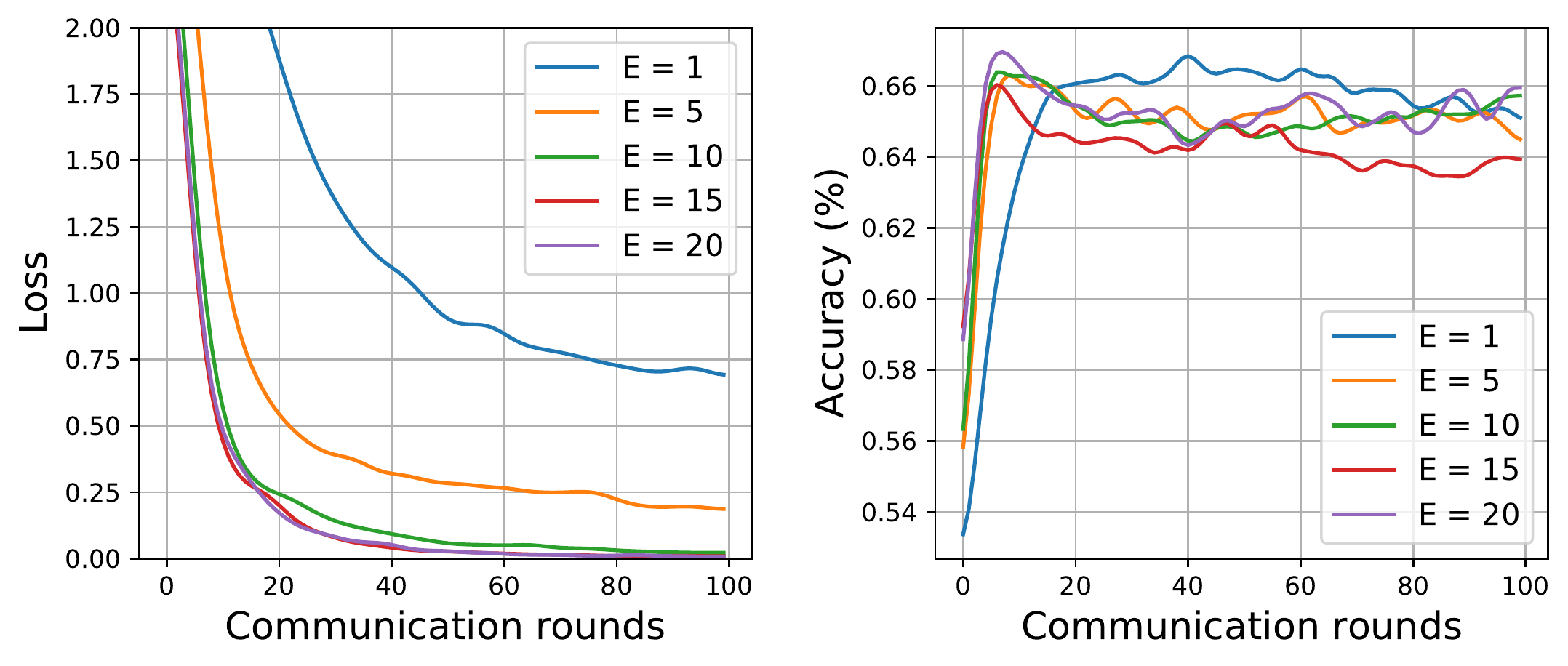}}}\\
\subfloat[ADNI]{\scalebox{0.4}{\includegraphics{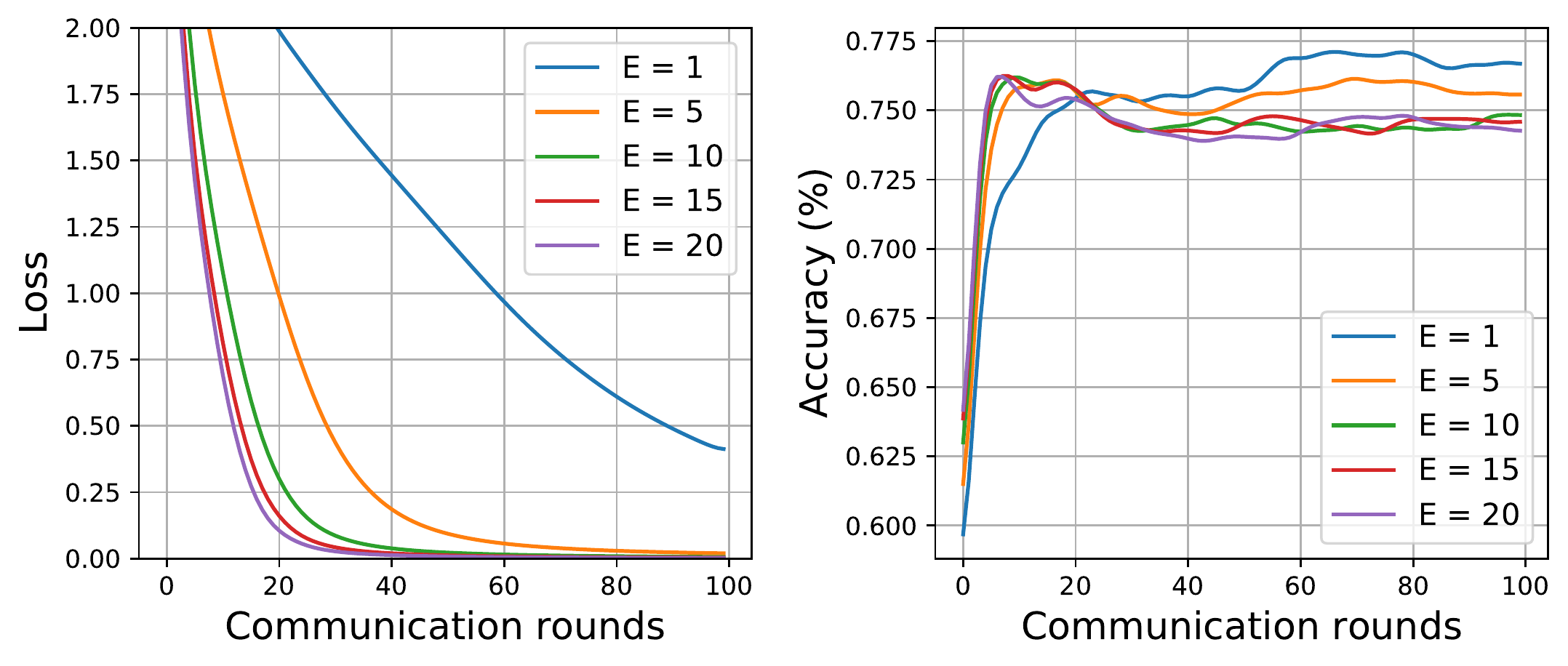}}}
\caption{The loss curve and the accuracy curve on the server with different local updating epochs (\ie $E = [1, 5, 10, 15, 20]$) on two datasets.}
\vspace{-2mm}
\label{fig53}
\end{figure}

We investigate the sensitivity of local updating epochs $E$ in \ours{}.
We evaluate the training convergence and test accuracy with different values of local updating epochs $E \in \{1,5,10,15,20\}$. 
\revision{As shown in Fig.~\ref{fig53}, we can observed that the loss is decreasing with the increasing communication rounds of training process, and thus our method has a good convergence property. Moreover, training large local updating epochs requires fewer global communication rounds to converge, while resulting in lower model performance, especially in the ADNI dataset.}\revisionfoot{\footnote{Reviewer 1, Q3}}

\section{Conclusion}
In this work,  we have proposed a new FL framework for distributed local population network analysis.
To tackle the realistic but ignored issue of incomplete information in local networks, we designed a federated network inpainting module, where a missing node generator allows each institution to generate missing nodes and edges. We train the missing node generator and GCN-based node classification model with federation. Extensive experiments on public datasets demonstrate that our method obtains state-of-the-art performance.  Our future work will address the data heterogeneity issue to further improve performance.

\section{Acknowledgment}

This work was partially supported by the National Natural Science Foundation of
China (Grant No. 61876046), Medico-Engineering Cooperation Funds from University of Electronic Science and Technology of China (No. ZYGX2022YGRH009
and ZYGX2022YGRH014), the Guangxi “Bagui” Teams for Innovation and
Research, China, and Natural Sciences and Engineering Research Council of Canada (GECR-2022-00430).

The authors are sincerely thankful for the support by NVIDIA and Microsoft Azure. We thank Dr. Qingyu Zhao at Stanford University for the data collection and Dr. Renjie Liao at the University of British Columbia for insightful suggestions.

\normalem{

}


\begin{thebibliography}{10}
\providecommand{\url}[1]{#1}
\csname url@samestyle\endcsname
\providecommand{\newblock}{\relax}
\providecommand{\bibinfo}[2]{#2}
\providecommand{\BIBentrySTDinterwordspacing}{\spaceskip=0pt\relax}
\providecommand{\BIBentryALTinterwordstretchfactor}{4}
\providecommand{\BIBentryALTinterwordspacing}{\spaceskip=\fontdimen2\font plus
\BIBentryALTinterwordstretchfactor\fontdimen3\font minus
  \fontdimen4\font\relax}
\providecommand{\BIBforeignlanguage}[2]{{%
\expandafter\ifx\csname l@#1\endcsname\relax
\typeout{** WARNING: IEEEtran.bst: No hyphenation pattern has been}%
\typeout{** loaded for the language `#1'. Using the pattern for}%
\typeout{** the default language instead.}%
\else
\language=\csname l@#1\endcsname
\fi
#2}}
\providecommand{\BIBdecl}{\relax}
\BIBdecl

\bibitem{stevens2021ankyrins}
S.~R. Stevens and M.~N. Rasband, ``Ankyrins and neurological disease,''
  \emph{Current Opinion in Neurobiology}, vol.~69, pp. 51--57, 2021.

\bibitem{wee2016diagnosis}
C.-Y. Wee, P.-T. Yap, and D.~Shen, ``Diagnosis of autism spectrum disorders
  using temporally distinct resting-state functional connectivity networks,''
  \emph{CNS neuroscience \& therapeutics}, vol.~22, no.~3, pp. 212--219, 2016.

\bibitem{song2020classification}
X.~Song, A.~Elazab, and Y.~Zhang, ``Classification of mild cognitive impairment
  based on a combined high-order network and graph convolutional network,''
  \emph{Ieee Access}, vol.~8, pp. 42\,816--42\,827, 2020.

\bibitem{huang2020edge}
Y.~Huang and A.~C. Chung, ``Edge-variational graph convolutional networks for
  uncertainty-aware disease prediction,'' in \emph{International Conference on
  Medical Image Computing and Computer-Assisted Intervention}.\hskip 1em plus
  0.5em minus 0.4em\relax Springer, 2020, pp. 562--572.

\bibitem{song2019graph}
T.-A. Song, S.~R. Chowdhury, F.~Yang, H.~Jacobs, G.~El~Fakhri, Q.~Li,
  K.~Johnson, and J.~Dutta, ``Graph convolutional neural networks for
  alzheimer’s disease classification,'' in \emph{2019 IEEE 16th International
  Symposium on Biomedical Imaging (ISBI 2019)}.\hskip 1em plus 0.5em minus
  0.4em\relax IEEE, 2019, pp. 414--417.

\bibitem{li2018brain}
H.~Li and Y.~Fan, ``Brain decoding from functional mri using long short-term
  memory recurrent neural networks,'' in \emph{International Conference on
  Medical Image Computing and Computer-Assisted Intervention}.\hskip 1em plus
  0.5em minus 0.4em\relax Springer, 2018, pp. 320--328.

\bibitem{spasov2019parameter}
S.~Spasov, L.~Passamonti, A.~Duggento, P.~Lio, N.~Toschi, A.~D.~N. Initiative
  \emph{et~al.}, ``A parameter-efficient deep learning approach to predict
  conversion from mild cognitive impairment to alzheimer's disease,''
  \emph{Neuroimage}, vol. 189, pp. 276--287, 2019.

\bibitem{dvornek2019jointly}
N.~C. Dvornek, X.~Li, J.~Zhuang, and J.~S. Duncan, ``Jointly discriminative and
  generative recurrent neural networks for learning from fmri,'' in
  \emph{International Workshop on Machine Learning in Medical Imaging}.\hskip
  1em plus 0.5em minus 0.4em\relax Springer, 2019, pp. 382--390.

\bibitem{sudha2021recurrrent}
V.~P. Sudha and M.~Vijaya, ``Recurrrent neural network based model for autism
  spectrum disorder prediction using codon encoding,'' \emph{Journal of The
  Institution of Engineers (India): Series B}, pp. 1--7, 2021.

\bibitem{parisot2018disease}
S.~Parisot, S.~I. Ktena, E.~Ferrante, M.~Lee, R.~Guerrero, B.~Glocker, and
  D.~Rueckert, ``Disease prediction using graph convolutional networks:
  application to autism spectrum disorder and alzheimer’s disease,''
  \emph{Medical image analysis}, vol.~48, pp. 117--130, 2018.

\bibitem{voigt2017eu}
P.~Voigt and A.~Von~dem Bussche, ``The eu general data protection regulation
  (gdpr),'' \emph{A Practical Guide, 1st Ed., Cham: Springer International
  Publishing}, vol.~10, p. 3152676, 2017.

\bibitem{act1996health}
A.~Act, ``Health insurance portability and accountability act of 1996,''
  \emph{Public law}, vol. 104, p. 191, 1996.

\bibitem{li2020multi}
X.~Li, Y.~Gu, N.~Dvornek, L.~H. Staib, P.~Ventola, and J.~S. Duncan,
  ``Multi-site fmri analysis using privacy-preserving federated learning and
  domain adaptation: Abide results,'' \emph{Medical Image Analysis}, vol.~65,
  p. 101765, 2020.

\bibitem{dayan2021federated}
I.~Dayan, H.~R. Roth, A.~Zhong, A.~Harouni, A.~Gentili, A.~Z. Abidin, A.~Liu,
  A.~B. Costa, B.~J. Wood, C.-S. Tsai \emph{et~al.}, ``Federated learning for
  predicting clinical outcomes in patients with covid-19,'' \emph{Nature
  medicine}, vol.~27, no.~10, pp. 1735--1743, 2021.

\bibitem{yang2021flop}
Q.~Yang, J.~Zhang, W.~Hao, G.~Spell, and L.~Carin, ``Flop: Federated learning
  on medical datasets using partial networks,'' \emph{arXiv preprint
  arXiv:2102.05218}, 2021.

\bibitem{astillo2021federated}
P.~V. Astillo, D.~G. Duguma, H.~Park, J.~Kim, B.~Kim, and I.~You, ``Federated
  intelligence of anomaly detection agent in iotmd-enabled diabetes management
  control system,'' \emph{Future Generation Computer Systems}, 2021.

\bibitem{he2021fedgraphnn}
C.~He, K.~Balasubramanian, E.~Ceyani, C.~Yang, H.~Xie, L.~Sun, L.~He, L.~Yang,
  P.~S. Yu, Y.~Rong \emph{et~al.}, ``Fedgraphnn: A federated learning system
  and benchmark for graph neural networks,'' \emph{arXiv preprint
  arXiv:2104.07145}, 2021.

\bibitem{hanneman2005introduction}
R.~A. Hanneman and M.~Riddle, ``Introduction to social network methods,'' 2005.

\bibitem{zhang2021subgraph}
K.~Zhang, C.~Yang, X.~Li, L.~Sun, and S.~M. Yiu, ``Subgraph federated learning
  with missing neighbor generation,'' \emph{arXiv preprint arXiv:2106.13430},
  2021.

\bibitem{zhu2020transfer}
Q.~Zhu, Y.~Xu, H.~Wang, C.~Zhang, J.~Han, and C.~Yang, ``Transfer learning of
  graph neural networks with ego-graph information maximization,'' \emph{arXiv
  preprint arXiv:2009.05204}, 2021.

\bibitem{taguchi2021graph}
H.~Taguchi, X.~Liu, and T.~Murata, ``Graph convolutional networks for graphs
  containing missing features,'' \emph{Future Generation Computer Systems},
  vol. 117, pp. 155--168, 2021.

\bibitem{mcmahan2017communication}
B.~McMahan, E.~Moore, D.~Ramage, S.~Hampson, and B.~A. y~Arcas,
  ``Communication-efficient learning of deep networks from decentralized
  data,'' in \emph{Artificial intelligence and statistics}.\hskip 1em plus
  0.5em minus 0.4em\relax PMLR, 2017, pp. 1273--1282.

\bibitem{miyato2018spectral}
T.~Miyato, T.~Kataoka, M.~Koyama, and Y.~Yoshida, ``Spectral normalization for
  generative adversarial networks,'' \emph{arXiv preprint arXiv:1802.05957},
  2018.

\bibitem{jiang2020hi}
H.~Jiang, P.~Cao, M.~Xu, J.~Yang, and O.~Zaiane, ``Hi-gcn: A hierarchical graph
  convolution network for graph embedding learning of brain network and brain
  disorders prediction,'' \emph{Computers in Biology and Medicine}, vol. 127,
  p. 104096, 2020.

\bibitem{yi2019not}
J.~Yi, J.~Lee, K.~J. Kim, S.~J. Hwang, and E.~Yang, ``Why not to use zero
  imputation? correcting sparsity bias in training neural networks,''
  \emph{arXiv preprint arXiv:1906.00150}, 2019.

\bibitem{luo2018multivariate}
Y.~Luo, X.~Cai, Y.~Zhang, J.~Xu, and X.~Yuan, ``Multivariate time series
  imputation with generative adversarial networks,'' in \emph{Proceedings of
  the 32nd International Conference on Neural Information Processing Systems},
  2018, pp. 1603--1614.

\bibitem{abraham2017deriving}
A.~Abraham, M.~P. Milham, A.~Di~Martino, R.~C. Craddock, D.~Samaras,
  B.~Thirion, and G.~Varoquaux, ``Deriving reproducible biomarkers from
  multi-site resting-state data: An autism-based example,'' \emph{NeuroImage},
  vol. 147, pp. 736--745, 2017.

\bibitem{zhu2022interpretable}
Y.~Zhu, J.~Ma, C.~Yuan, and X.~Zhu, ``Interpretable learning based dynamic
  graph convolutional networks for alzheimer’s disease analysis,''
  \emph{Information Fusion}, vol.~77, pp. 53--61, 2022.

\bibitem{li2020review}
L.~Li, Y.~Fan, M.~Tse, and K.-Y. Lin, ``A review of applications in federated
  learning,'' \emph{Computers \& Industrial Engineering}, p. 106854, 2020.

\bibitem{li2019privacy}
W.~Li, F.~Milletar{\`\i}, D.~Xu, N.~Rieke, J.~Hancox, W.~Zhu, M.~Baust,
  Y.~Cheng, S.~Ourselin, M.~J. Cardoso \emph{et~al.}, ``Privacy-preserving
  federated brain tumour segmentation,'' in \emph{International workshop on
  machine learning in medical imaging}.\hskip 1em plus 0.5em minus 0.4em\relax
  Springer, 2019, pp. 133--141.

\bibitem{kipf2016semi}
T.~N. Kipf and M.~Welling, ``Semi-supervised classification with graph
  convolutional networks,'' \emph{arXiv preprint arXiv:1609.02907}, 2016.

\bibitem{hamilton2017inductive}
W.~L. Hamilton, R.~Ying, and J.~Leskovec, ``Inductive representation learning
  on large graphs,'' in \emph{NIPS}, 2017, pp. 1025--1035.

\bibitem{wu2020comprehensive}
Z.~Wu, S.~Pan, F.~Chen, G.~Long, C.~Zhang, and S.~Y. Philip, ``A comprehensive
  survey on graph neural networks,'' \emph{IEEE transactions on neural networks
  and learning systems}, vol.~32, no.~1, pp. 4--24, 2020.

\bibitem{zhao2020differentiable}
S.~Zhao, Z.~Liu, J.~Lin, J.-Y. Zhu, and S.~Han, ``Differentiable augmentation
  for data-efficient gan training,'' \emph{arXiv preprint arXiv:2006.10738},
  2020.

\bibitem{zhu2020deep}
L.~Zhu and S.~Han, ``Deep leakage from gradients,'' in \emph{Federated
  learning}.\hskip 1em plus 0.5em minus 0.4em\relax Springer, 2020, pp. 17--31.

\bibitem{dwork2014algorithmic}
C.~Dwork, A.~Roth \emph{et~al.}, ``The algorithmic foundations of differential
  privacy.'' \emph{Found. Trends Theor. Comput. Sci.}, vol.~9, no. 3-4, pp.
  211--407, 2014.

\bibitem{di2014autism}
A.~Di~Martino, C.-G. Yan, Q.~Li, E.~Denio, F.~X. Castellanos, K.~Alaerts, J.~S.
  Anderson, M.~Assaf, S.~Y. Bookheimer, M.~Dapretto \emph{et~al.}, ``The autism
  brain imaging data exchange: {T}owards a large-scale evaluation of the
  intrinsic brain architecture in autism,'' \emph{Molecular Psychiatry},
  vol.~19, no.~6, pp. 659--667, 2014.

\bibitem{timmurphy.org}
``Alzheimer’s disease neuroimaging initiative (adni),''
  \url{http://adni.loni.usc.edu}, 2003.

\bibitem{DPARSF}
C.~Yan and Y.~Zang, ``Dparsf: a matlab toolbox for "pipeline" data analysis of
  resting-state fmri,'' \emph{Frontiers in Systems Neuroscience}, vol.~4,
  p.~13, 2010.

\bibitem{sled1998nonparametric}
J.~G. Sled, A.~P. Zijdenbos, and A.~C. Evans, ``A nonparametric method for
  automatic correction of intensity nonuniformity in mri data,'' \emph{IEEE
  transactions on medical imaging}, vol.~17, no.~1, pp. 87--97, 1998.

\bibitem{fischl2012freesurfer}
B.~Fischl, ``Freesurfer,'' \emph{Neuroimage}, vol.~62, no.~2, pp. 774--781,
  2012.

\bibitem{baronzio1999proinflammatory}
G.~Baronzio, A.~Zambelli, D.~Comi, A.~Barlocco, A.~Baronzio, P.~Marchesi,
  A.~Gramaglia, E.~Castiglioni, A.~Mafezzoni, E.~Beviglia \emph{et~al.},
  ``Proinflammatory and regulatory cytokine levels in aids cachexia.'' \emph{In
  vivo (Athens, Greece)}, vol.~13, no.~6, pp. 499--502, 1999.

\end{thebibliography}
\end{document}